\documentclass[journal]{IEEEtran}
\ifCLASSINFOpdf
\else
\fi

\usepackage{cite}
\usepackage{amsmath,amssymb,amsfonts}
\usepackage{graphicx}
\usepackage{textcomp}
\usepackage{xcolor}
\usepackage{float}
\usepackage[ruled,vlined]{algorithm2e}
\usepackage{algpseudocode}
\usepackage{makecell}
\usepackage{caption}
\usepackage{subfigure}
\usepackage{mdwlist}
\usepackage{enumitem} 
\usepackage{mathrsfs}
\usepackage{booktabs}
\usepackage{mathtools}
\usepackage{flushend}

\IEEEoverridecommandlockouts

\begin{document}
%
\title{\huge Hardware Acceleration of Explainable Artificial Intelligence}



%


\author{Zhixin~Pan,~\IEEEmembership{Member,~IEEE,}
        and~Prabhat~Mishra,~\IEEEmembership{Fellow,~IEEE,}
\IEEEcompsocitemizethanks{\IEEEcompsocthanksitem Z. Pan and P. Mishra are with the Department of Computer \& Information Science \& Engineering, University of Florida, Gainesville, Florida, USA. Email: panzhixin@ufl.edu}
\thanks{This work was partially supported by the NSF grant CCF-1908131.}
}


\maketitle

\begin{abstract}
Machine learning (ML) is successful in achieving human-level artificial intelligence in various fields. However, it lacks the ability to explain an outcome due to its black-box nature. While recent efforts on explainable AI (XAI) has received significant attention, most of the existing solutions are not applicable in real-time systems since they map interpretability as an optimization problem, which leads to numerous iterations of time-consuming complex computations.
Although there are existing hardware-based acceleration framework for XAI, they are implemented through FPGA and designed for specific tasks, leading to expensive cost and lack of flexibility. In this paper, we propose a simple yet efficient framework to accelerate various XAI algorithms with existing hardware accelerators. Specifically, this paper makes three important contributions. (1) The proposed method is the first attempt in exploring the effectiveness of Tensor Processing Unit (TPU) to accelerate XAI. (2) Our proposed solution explores the close relationship between several existing XAI algorithms with matrix computations, and exploits the synergy between convolution and Fourier transform, which takes full advantage of TPU's inherent ability in accelerating matrix computations. (3) Our proposed approach can lead to real-time outcome interpretation. Extensive experimental evaluation demonstrates that proposed approach deployed on TPU can provide drastic improvement in interpretation time (39x on average) as well as energy efficiency  (69x on average) compared to existing acceleration techniques. 


\end{abstract}


%
\IEEEpeerreviewmaketitle

\section{Introduction}




Machine learning (ML) techniques powered by deep neural networks (DNNs) are pervasive across various application domains. Recent advances in ML algorithms have enabled promising performance with outstanding flexibility and generalization to achieve human-level artificial intelligence. However, most of the existing ML methods are not able to interpret the outcome (e.g., explain its prediction) since it produces the outcome based on computations inside a ``black-box''. This lack of transparency severely limits the applicability of ML. 
In many applications, a designer or user can act judiciously if an ML algorithm can provide an outcome as well as an  interpretation of the outcome. For example, during malware detection, it is important to know whether a software is malicious or benign as well as the rationale for such a classification. Moreover, the interpretation of the results is crucial to enable the localization (e.g., clock cycle and specific reason) of the malicious activity in a malware \cite{iccd20}.  
 
\begin{figure}[htbp]
\centering
\vspace{-0.1in}
\includegraphics[scale=0.44]{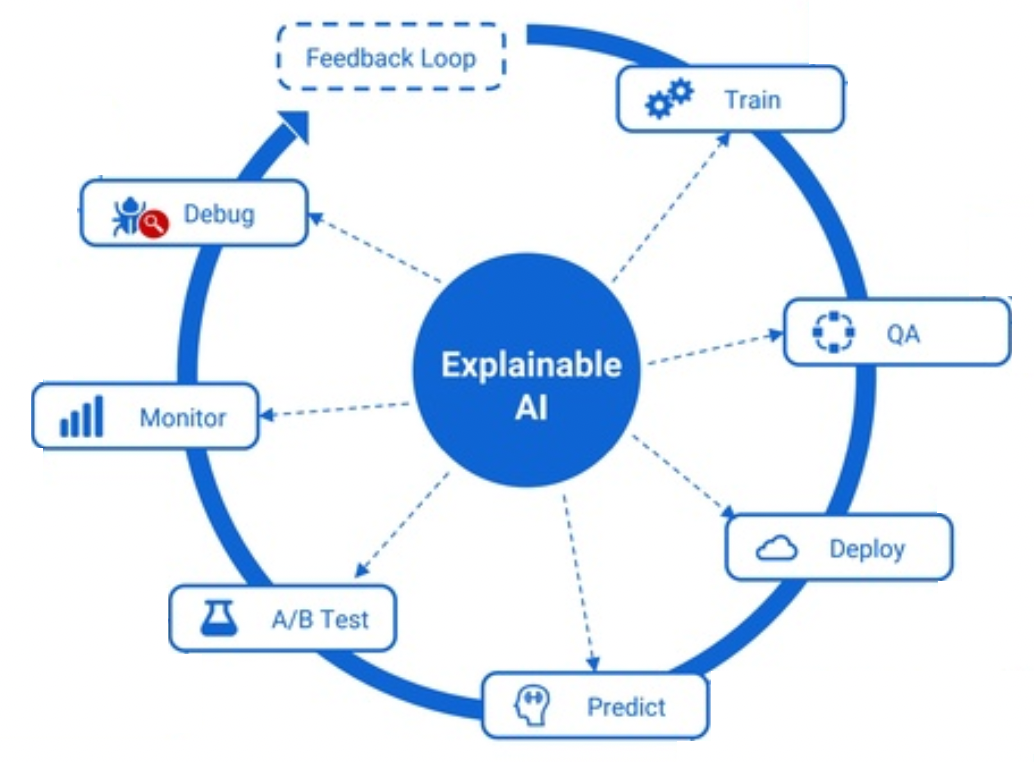}
\vspace{-0.1in}
\caption{Wide adoption of explainable machine learning.}
\label{fig:expAI}
\end{figure}

Explainable AI (XAI) is a promising direction to enable outcome interpretation. There are a large number of existing efforts in the area of XAI ~\cite{EML,smilkov2017smoothgrad,li2020does}. By providing interpretation of input-output mapping and clues for importance ranking of input features, XAI acts like another supervisor to guide the learning process and bring extra information to users. XAI can be adopted in many application domains as shown in Figure~\ref{fig:expAI}. For example, during the training of a facial recognition classifier, if we can obtain information about which region in the face distinguishes the target from the others, the corresponding weight can be adjusted to emphasize features collected from that region. 

Due to the inherent inefficiency in XAI algorithms, they are not applicable in real-time systems. These algorithms treat the explanation process as an extra procedure, and performs the interpretation outside the learning model (Figure~\ref{fig:modeldistill}), which makes them inefficient in practice. Specifically, it solves a complex optimization problem that consists of numerous iterations of time-consuming computations. As a result, such time-consuming interpretation is not suitable for time-sensitive applications with soft or hard deadlines. In soft real-time systems, such as multimedia and gaming devices, inefficient interpretation can lead to unacceptable Quality-of-Service (QoS). In hard real-time systems, such as safety-critical systems, missing task deadlines can lead to catastrophic consequences. While there are recent efforts in designing FPGA-based hardware accelerators for XAI algorithms, it requires customization of FPGA for the target XAI algorithm, which limits its applicability.   

In this paper, we propose an efficient framework to achieve fast XAI utilizing Tensor Processing Units (TPU). Specifically, we transform various XAI algorithms (model distillation, Shapley value analysis, and Integrated Gradient) into computation of matrix operations, and exploit the natural ability of TPU for fast and efficient parallel computation of matrix operations. TPU is developed to accelerate tensor computations in deep neural networks~\cite{TPU1,TPU2,TPU3,TPU4,TPU5}. It provides extremely high throughout and fast performance with low memory footprint, and is widely applied in accelerating general ML procedure. As a result, there is no need to implement new application-specific hardware components, and our proposed framework has a natural compatibility with any existing acceleration framework for general ML process. 


This paper makes the following major contributions:

\begin{enumerate}
    \item To the best of our knowledge, our proposed approach is the first attempt in exploring the effectiveness of Tensor Processing  Unit (TPU) in accelerating XAI. 
    \item We propose an efficient mechanism to convert three most widely-applied XAI algorithms (model distillation, Shapley value analysis, and integrated gradient) to linear algebra computation. As a result, it can exploit the inherent advantages of hardware accelerators in computing ultra-fast matrix operations. 
    \item Experiments using two popular ML models demonstrate that our proposed approach can achieve real-time explainable machine learning by providing fast outcome interpretation, with significant improvement compared with existing state-of-the-art hardware accelerators.
\end{enumerate}

The rest of this paper is organized as follows. Section~\ref{relwork} surveys related efforts. Section~\ref{proposed} describes our proposed method for accelerating explainable AI algorithms using TPU/GPU-based architectures. Section~\ref{exp} presents experimental results. Finally, Section~\ref{conclude} concludes the paper.

\section{Background and Related Work}
\label{relwork}

The proposed work has two major components: hardware acceleration and explainable artificial intelligence (XAI). We first describe hardware accelerators for machine learning algorithms. Next, we discuss three popular XAI models: model distillation, Shapley value analysis, and integrated gradient. 

\subsection{Hardware Accelerators for Machine Learning Algorithms}
\label{sec:gputpu}

The objective of hardware acceleration is to utilize computer hardware to speed up artificial intelligence applications faster than what would be possible with a software running on a general-purpose central processing unit (CPU). 

\begin{figure}[htbp]
\centering
\vspace{-0.1in}
\includegraphics[scale=0.32]{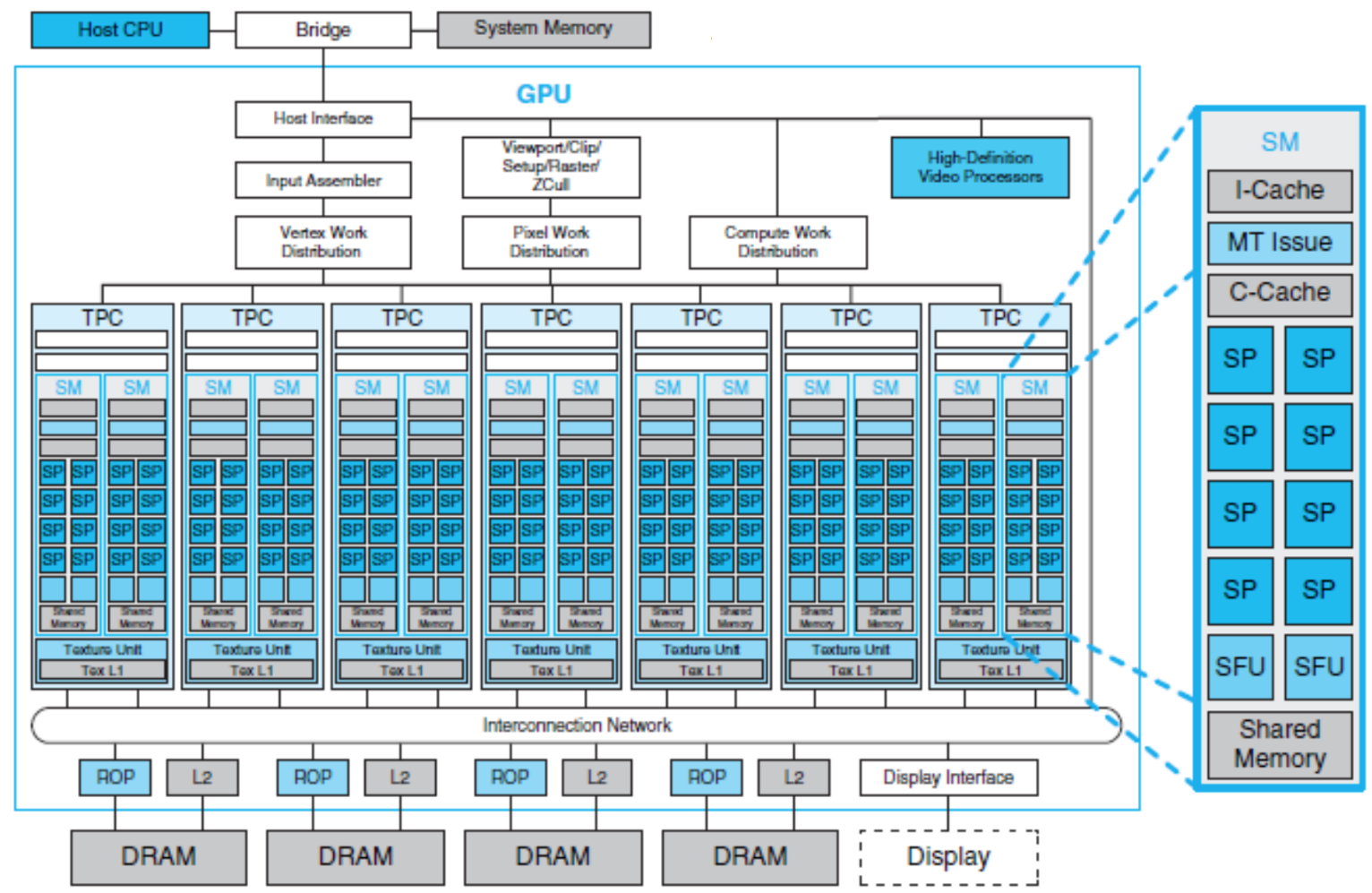}
\caption{An example NVDIA GPU architecture with a large number of processor cores organized in multiple clusters.}
\label{fig:structure}
\vspace{-0.1in}
\end{figure}

\vspace{0.1in}
\noindent \underline{\textit{GPU-based Acceleration of ML models}}:
Graphical Processing Units (GPUs) is a general-purpose solution for accelerating ML models. Fundamentally, CPU and GPU's performance in ML are significantly different because they have different design goals, i.e., they are aimed at two different application scenarios. The CPU needs strong versatility to handle various data types, and at the same time it requires logical judgment and introduces a large number of branches for interruption processing. These all make the internal structure of the CPU extremely complicated. In contrast, the GPU deals with a highly unified, independent, large-scale data and a pure computing environment that does not need to be interrupted. 
GPU uses a large number of computing units and long pipelines as shown in Figure~\ref{fig:structure}. GPU provides hundreds of simpler cores (vs a few complex), allows thousands of threads to run in parallel (vs single thread performance optimization), and allocates most of the die surface for simple integer and floating point operations (vs more complex operations). Therefore, GPU is more powerful than CPU for handling large number of parallel computing tasks, including matrix multiplication and convolution operations in deep learning algorithms. 

\vspace{0.1in}
\noindent \underline{\textit{FPGA-based Acceleration of ML models}}:
Another widely-adopted hardware accelerator is Field-Programmable Gate Array (FPGA). The advantage of this kind of specific purpose integrated circuit compared to a general-purpose one is its flexibility: after manufacturing it can be programmed to implement virtually any possible design and be more suited to the application on hand compared to a GPU. FPGA is widely used for fast prototyping as well as acceleration of diverse algorithms. FPGAs provide reconfigurability and faster on-chip memory, resulting in higher computation capability. This on-chip memory reduces bottlenecks caused by external memory access and reduces the cost and power required for high memory bandwidth solutions compared to CPU-based software implementation. However, FPGA-based accelerators introduce significant cost for designing and manufacturing FPGAs customized for specific algorithms. Specifically, ML tasks involves multiple types of models that may require multiple types of FPGAs to implement them, making it infeasible for large-scale applications.

\begin{figure}[htbp]
\centering
\vspace{-0.1in}
\includegraphics[scale=0.29]{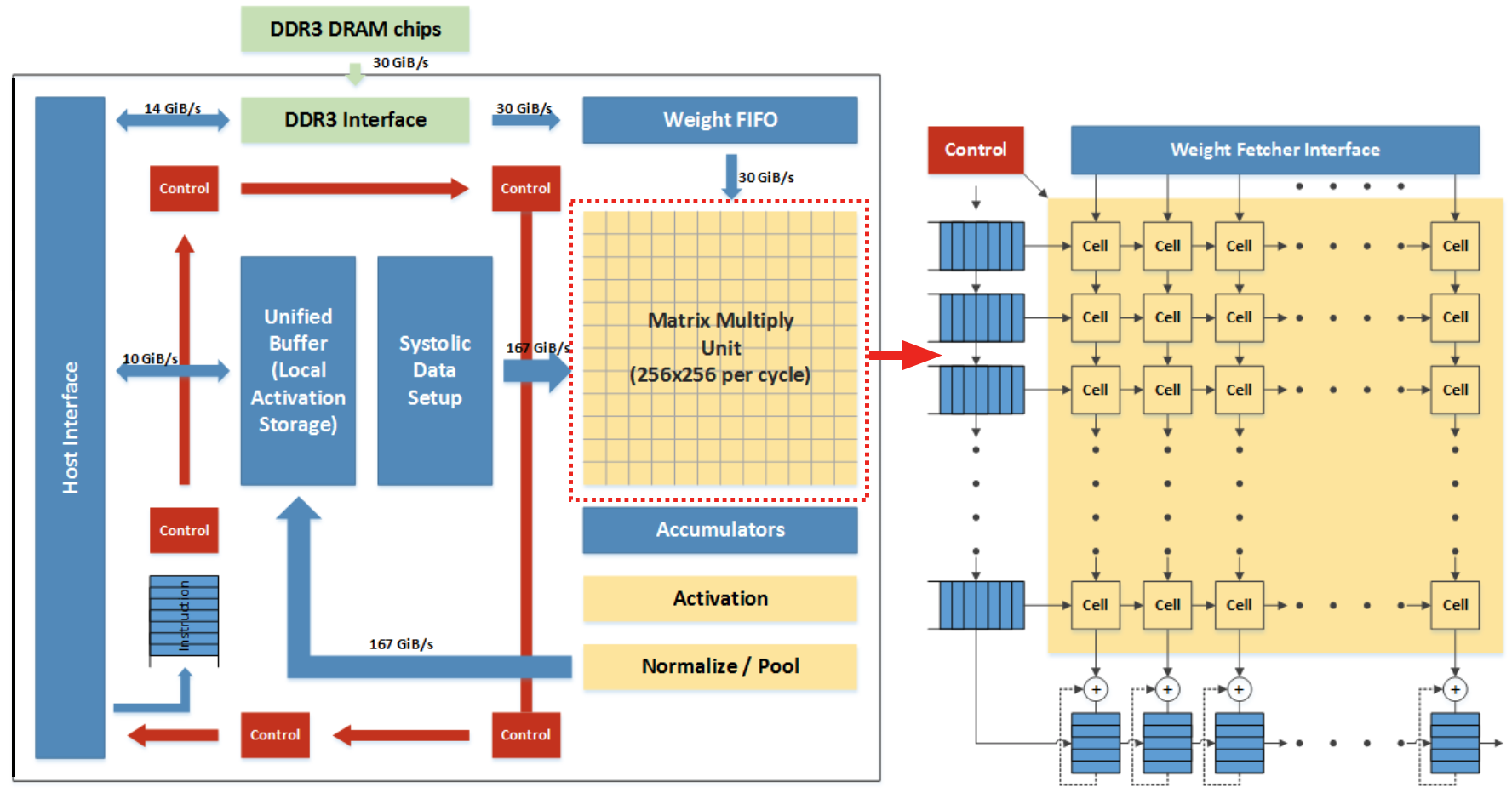}
\vspace{-0.2in}
\caption{An example TPU architecture where Matrix Multiply Unit (MXU) is implemented by systolic array.}
\label{fig:structureTPU}
\vspace{-0.1in}
\end{figure}

\vspace{0.1in}
\noindent \underline{\textit{TPU-based Acceleration of ML models}}:
Tensor Processing Unit (TPU) is a special case of domain-specific hardware for accelerating computation process of deep learning models. Comparing with GPU and other FPGAs, there are two fundamental reasons for the superior performance of TPUs:  \textit{quantization} and \textit{systolic array}~\cite{SysArry}. Quantization is the first step of optimization, which uses 8-bit integers to approximate 16-bit or 32-bit floating-point numbers. This can reduce the required memory capacity and computing resources. Systolic array is a major contributor to TPU's efficiency due to its natural compatibility with matrix manipulation coupled with the fact that computation in neural networks can be represented as matrix operations. Figure~\ref{fig:structureTPU} shows an overview of the simplified structure of a TPU.

Informally, the core of the entire TPU is the Matrix Multiply Unit, which is a 256$\times$256 systolic array composed of multiple computation cells. Each cell receives a weight parameter along with an input signal at a time, and performs accumulation of their products. Once all weights and input signals are propagated to neighboring cells, top to bottom and left to right respectively, it immediately starts next round of computations. As a result, the entire matrix multiplication can be  completed by the collaboration of all computation cells. The systolic array of MXU contains 256 × 256 = 65,536 ALUs, which means that the TPU can process 65,536 8-bit integer multiplications and additions per cycle. Due to the systolic architecture, input data can be reused for multiple times. Therefore, it can achieve higher throughput while consuming less memory bandwidth. 

\vspace{-0.1in}
\subsection{Explainable AI using Model Distillation}
The demand for explainable AI has been steadily increasing ever since ML algorithms were widely adopted in many fields, especially in security domains. In this section, we briefly introduce explainable AI, and then discuss one widely adopted explanation technique. In general, explainable AI seeks to provide interpretable explanation for the results of machine learning model. Specifically, given an input instance $\bf x$ and a model $M$, the classifier will generate a corresponding output $\bf y$ for $\bf x$ during the testing time. Explanation techniques then aim to illustrate why instance $\bf x$ is transformed into $\bf y$. This often involves identifying a set of important features that make key contributions to the forward pass of model $M$. If the selected features are interpretable by human analysts, then these features can offer an “explanation”. 

\begin{figure}[htbp]
\centering
\vspace{-0.1in}
\includegraphics[scale=0.3]{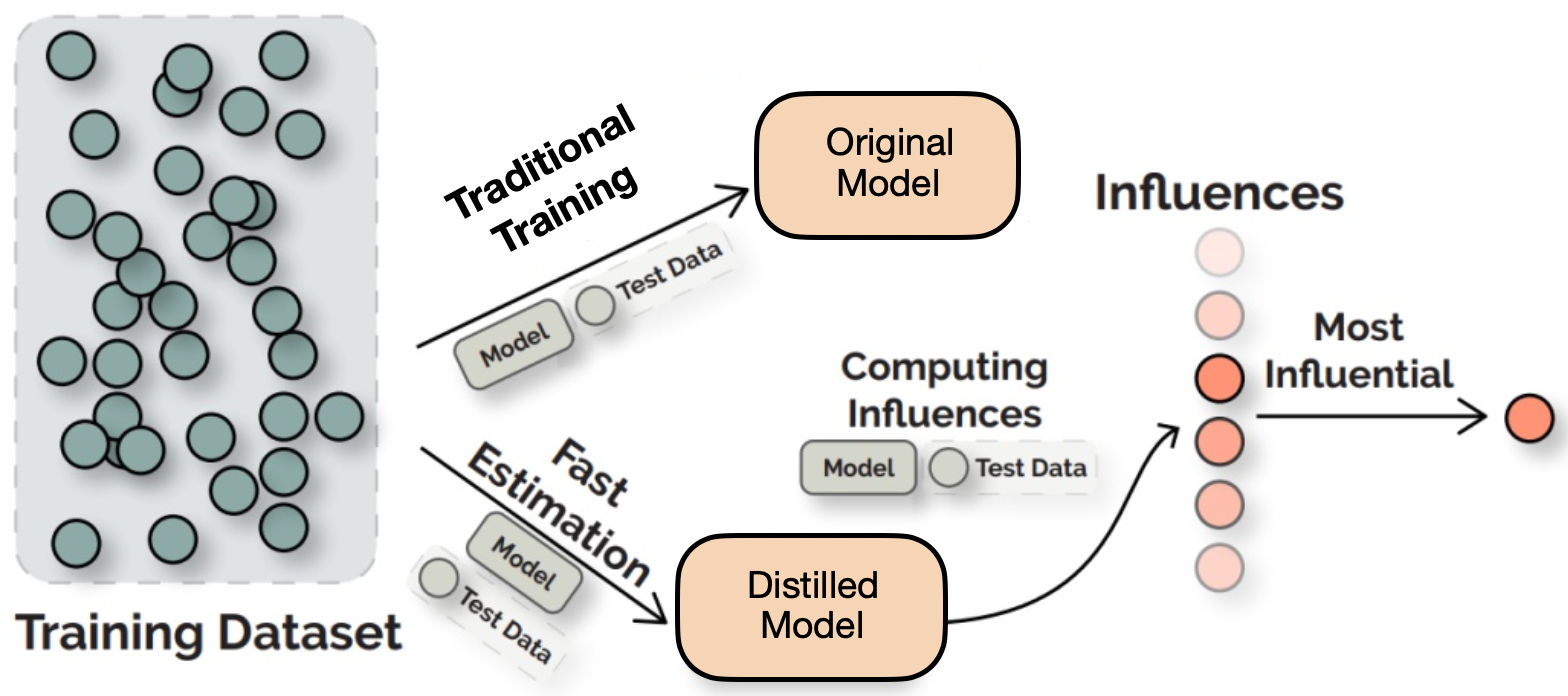}
\caption{The general process of explainable machine learning. By providing influences of input features, the model can be interpreted in a human-understandable way.}
\label{fig:modeldistill}
\vspace{-0.1in}
\end{figure}

One of the explanation methods exploit the concept of ``model distillation''~\cite{ModelDistillation}. The basic idea of model distillation is that it develops a separate model called as ``distilled model'' to be an approximation of the input-output behavior of the target machine learning model. This distilled model, denoted as $M^*$, is usually chosen to be inherently explainable by which a user is able to identify the decision rules or input features influencing the outputs. This is depicted in Figure~\ref{fig:modeldistill} and in reality, model distillation is composed of three major steps.

\vspace{0.05in}
\noindent \underline{\textit{Model Specification}}: The type of distilled model has to be specified at first. This often involves a challenging trade-off between transparency and expression ability. A complex model can offer better performance in mimicking the behavior of original model $M$. However, increasing complexity also leads to the inevitable drop of model transparency, where the distilled model itself becomes hard to explain, and vice versa.

\vspace{0.05in}
\noindent \underline{\textit{Model Computation}}: Once the type of distilled model $M^*$ is determined and corresponding input-output pairs are provided, the model computation task aims at searching for optimal parameters $\theta$ for $M$ using Equation~\ref{eqn:theta}, which is an optimization problem.  
    \vspace{-0.1in}
    \begin{equation}
    \theta = \arg\min\limits_{\bf {\theta}}||M^*({\bf x}) -\textbf{y}|| 
    \label{eqn:theta}
    \vspace{-0.1in} 
    \end{equation}
    
\vspace{0.1in}
\noindent \underline{\textit{Outcome Interpretation}}: Based on the computed distilled model, the explanation boils down to measuring the contribution of each input feature in producing the classification result.  For instance, assume a linear regression model is applied which can always be expressed as a polynomial. Then by sorting the terms with amplitude of coefficients, we have access to many crucial information including the input features it found to be most discriminatory, and features or output correlations relevant for classification. Notice if we choose linear regression in the previous step, the entire model distillation process degenerates to the ``Saliency Map'' method~\cite{simonyan2013deep}. 

Note that interpreting the distilled model may not provide us with deep insights into the internal representation of the ML model, or demonstrate anything about the model's learning process, but it can at least provide insight into the correlations to  explain how the ML model makes a decision. 

\subsection{Explainable AI using Shapley Values}


The key idea of Shpaley analysis is similar to model distillation. Specifically, it aims to illustrate what is the major reason for model transferring certain input into its prediction, guided by identifying the important features that make key contributions to the forward pass of the model. 

The concept of Shapley values (SHAP) is borrowed from the cooperative game theory~\cite{roth1988shapley}. It is used to fairly attribute a player’s contribution to the end result of a game. SHAP captures the marginal contribution of each player to the final result. Formally, we can calculate the marginal contribution of the $i$-th player in the game by:
\begin{equation}
\phi_i = \sum\limits_{S \subseteq N/\{i\} } \frac{|S|!(M - |S| -1)!}{M!} [f_x(S \cup \{i\}) -f_x(S)]    
\end{equation}

where the total number of players is $|M|$. $S$ represents any subset of players that does not include the $i$-th player, and $f_x(\cdot)$ represents the function to give the game result for the subset S.
Intuitively, SHAP is a weighted average payoff gain that player $i$ provides if added into every possible coalitions without $i$. 

\begin{figure}[htp]
    \centering 
    \vspace{-0.1in}
    \includegraphics[scale = 0.47]{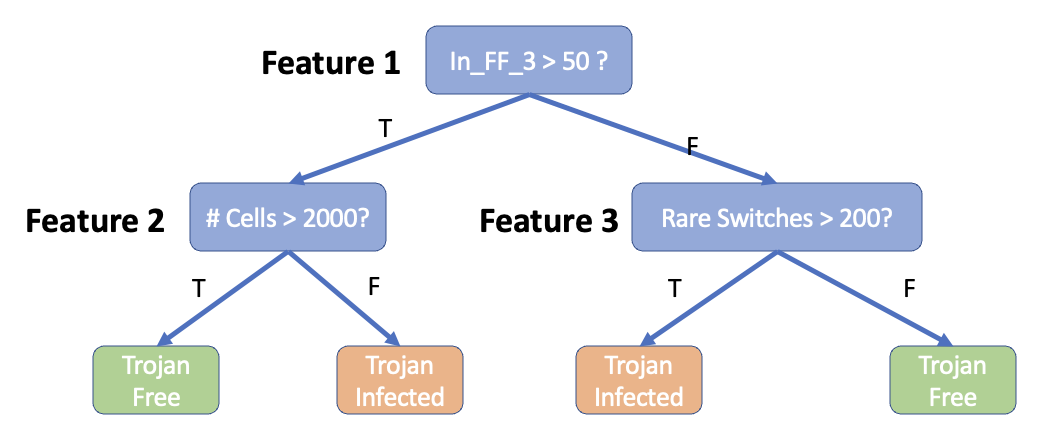}
    \vspace{-0.1in}
    \caption{Example decision tree that classifies using 3 features}
    \label{fig:dt}
    \vspace{-0.1in}
\end{figure}

To apply SHAP in ML tasks, we can assume features as the `players' in a cooperative game. SHAP is a local feature attribution technique that explains every prediction from the model as a summation of each individual feature contributions. Assume a decision tree is built with 3 different features for hardware Trojan (HT) detection as shown in Figure~\ref{fig:dt}. To compute SHAP values, we start with a null model without any independent features. Next, we compute the payoff gain as each feature is added to this model in a sequence. Finally, we compute average over all possible sequences. Since we have 3 independent variables here, we have to consider 3!=6 sequences. Specifically, the computation process for the SHAP value of the first feature is presented in Table~\ref{tab:mc}.

\begin{table}[h]
    \centering
    \small
            \vspace{-0.1in}
        \caption{Marginal contributions of the first feature.}
        \vspace{-0.05in}
    \begin{tabular}{|c|c|}
    \hline
    Sequences & Marginal Contributions\\
    \hline
    1,2,3     &  $\mathcal{L}(\{1\}) - \mathcal{L}(\emptyset)$\\
    1,3,2     & $\mathcal{L}(\{1\}) - \mathcal{L}(\emptyset)$\\
    2,1,3     & $\mathcal{L}(\{1,2\}) - \mathcal{L}(\{2\})$\\
    2,3,1     & $\mathcal{L}(\{1,2,3\}) - \mathcal{L}(\{2,3\})$\\
    3,1,2     & $\mathcal{L}(\{1,3\}) - \mathcal{L}(\{3\})$\\
    3,2,1     & $\mathcal{L}(\{1,2,3\}) - \mathcal{L}(\{3,2\})$\\
    \hline
    \end{tabular}
    \label{tab:mc}
    \vspace{-0.1in}
\end{table}

Here, $\mathcal{L}$ is the loss function. The loss function serves as the `score' function to indicate how much payoff currently we have by applying existing features. For example, in the first row, the sequence is $1,2,3$, meaning we sequentially add the first, second, and the third features into consideration for classification. Here, $\emptyset$ stands for the model without considering any features, which in our case is a random guess classifier, and $\mathcal{L}(\emptyset)$ is the corresponding loss. Then by adding the first feature into the scenario, we use $\{1\}$ to represent the dummy model that only uses this feature to perform prediction. We again compute the loss $\mathcal{L}(\{1\})$. $\mathcal{L}(\{1\}) - \mathcal{L}(\emptyset)$ is the marginal contribution of the first feature for this specific sequence. We obtain the SHAP value for the first feature by computing the marginal contributions of all 6 sequences and taking the average. Similar computations happen for the other features.


\vspace{-0.1in}
\subsection{Explainable AI using Integrated Gradient (IG)}
\label{sec:relIG}
Integrated Gradients (IG) was proposed by Sundararajan et al.~\cite{sundararajan2017axiomatic} as an alternate mechanism to explain ML models. The equation to compute the IG attribution for an input record $x$ and a baseline record $x’$ is as follows:
\[
IG_i(x) \coloneqq (x_i - {x_i}') \times \int\limits^1_{\alpha = 0} \frac{\partial F(x'+ \alpha \times (x - x'))}{\partial x_i} d\alpha
\]

where $F: \mathcal{R}^n \rightarrow [0,1]$ represents the ML model, $\frac{\partial F(x)}{\partial x_i}$ is the gradient of $F(x)$ in the $i$-th dimension. Informally, IG defines the attribution of the $i$-th feature $x_i$ from input as the integral of the straight path from baseline $x'$ to input $x$.


Compared to other XAI algorithms, IG satisfies many desirable axioms outlined in~\cite{sundararajan2017axiomatic}. We specifically highlight two of those axioms that are particularly important for an explanation method:

\begin{itemize}
    \item The completeness axiom: Given $x$ and a baseline $x’$, the attributions of $x$ add up to the difference between the output of F at the input x and the baseline x.
    \item The sensitivity axiom: For every input and baseline that differ in one feature but have different predictions,   the differing feature should be given a non-zero attribution.
\end{itemize}

\subsection{Related Work on Hardware Acceleration}
Both TPU and GPU architectures are widely used in accelerating diverse applications. For example, in ~\cite{sidea2021optimal}, the author applied GPU-based accelerator to solve the optimal scheduling problem fast to achieve efficient energy storage in distributed systems. Similarly, many computationally expensive mathematical algorithms can be accelerated by TPU/GPU implementations~\cite{nadim2021efficient, wei2021accelerating, prasad2019improving,kim2012time}. The popularity of AI algorithms in recent years has led to many research efforts in hardware acceleration of machine learning algorithms. In~\cite{yazdanbakhsh2021evaluation}, TPU was utilized on various \textit{Convolutional Neural Networks} (CNN) to demonstrate its advantage in performing convolution operations. This advantage was further exploited in ML-based tasks like image reconstruction~\cite{lu2020accelerating} and automatic path-finding~\cite{sengupta2020high}.

Hardware-based solutions are also recently applied for accelerating XAI. In~\cite{zhou2022model}, the author proposed a principled hardware architecture for high performance XAI applied on Graph Convolutional Networks (GNNs) using FPGA. In ~\cite{mitchell2022gputreeshap}, the author proposed GPUTreeShap, a reformulated TreeShap algorithm suitable for massively parallel computation on GPU. In~\cite{bhat2022gradient}, a gradient backpropagation based feature attribution algorithm is proposed to enable XAI, and the entire framework is deployed on FPGA to achieve execution-on-the-edge. The above approaches require application-specific customization of FPGA, which limits their applicability across XAI algorithms. Our proposed TPU-based acceleration maps an explainability procedure into matrix computations, and therefore, it is applicable across diverse explainable AI algorithms.


\section{Hardware Acceleration of Explainable AI}\label{proposed}

\begin{figure}[htbp]
\centering
\vspace{-0.1in}
\includegraphics[scale=0.41]{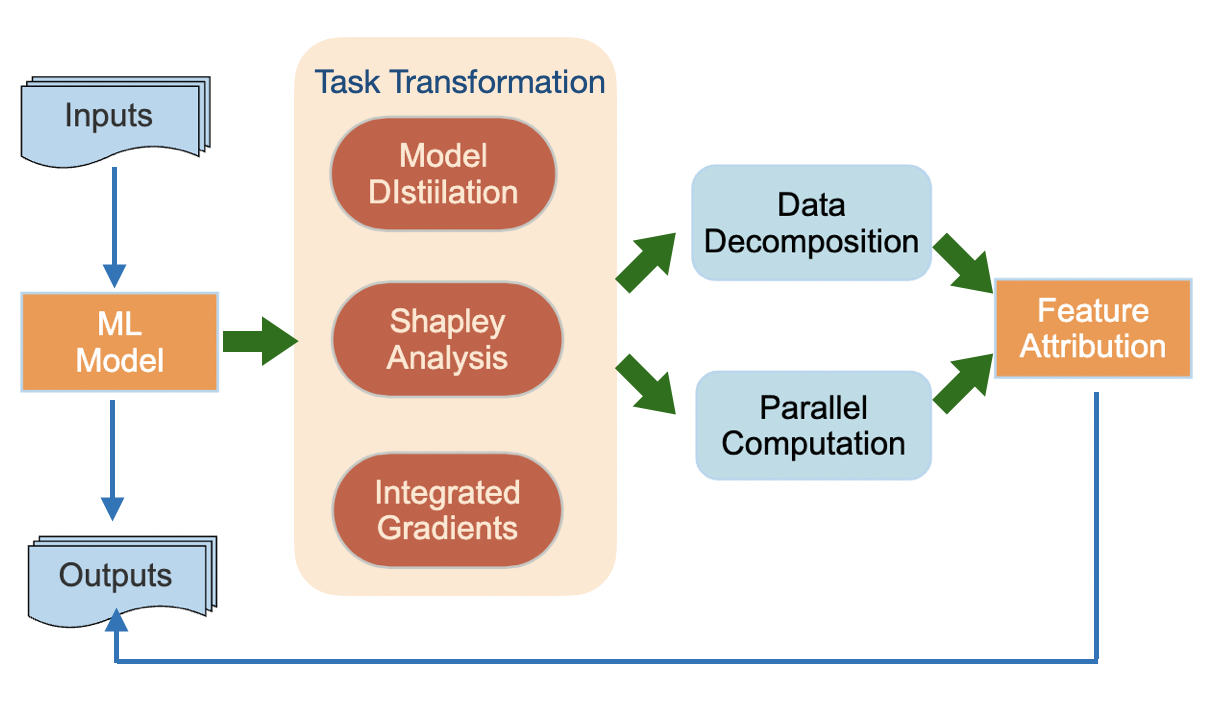}
\vspace{-0.2in}
\caption{Our proposed hardware acceleration framework consists of three major activities: task transformation, data decomposition, and parallel computation.}
\label{fig:overview}
\vspace{-0.1in}
\end{figure}

Figure~\ref{fig:overview} shows an overview of our proposed framework for hardware acceleration of explainable machine learning. For a specific ML task, we apply traditional training scheme to construct a well-trained model with respective input-output dataset. Then we apply various XAI algorithms (model distillation, Shapley analysis, and integrated gradients) to provide feature attributions for explaining target model's behavior. Specifically, these XAI algorithms are adjusted to accommodate TPU through the following three steps. First, we perform \textit{task transformation} to map the original XAI algorithms to matrix computations (Section~\ref{transMD}, Section~\ref{transSP}, and Section~\ref{transIG}). Next, we develop two synergistic activities to accelerate the computation procedure of Fourier transform. The first activity performs \textit{data decomposition} (Section~\ref{decomp}), where the complete computing task is split into multiple sub-tasks, and each sub-task can be executed by a TPU core without requiring any data exchange between cores (sub-tasks). The other one fully exploits hardware accelerators' inherent ability in \textit{parallel computation} (Section~\ref{paral}) to process multiple input-output pairs concurrently. Simultaneous execution of these two activities can provide significant improvement in acceleration  efficiency, which is demonstrated in our experimental evaluation (Section~\ref{exp}).

\subsection{Transformation of Model Distillation}
\label{transMD}
Model distillation consists of the following three steps: model specification, model computation and outcome interpretation. In this section, we discuss how to convert model distillation task into a matrix computation. 



\vspace{0.1in}
\textit{\textbf{Model Specification:}} To make the distilled model useful, two crucial requirements need to be satisfied.
\begin{itemize}
    \item \textit{Simplicity:} The distilled model should be lightweight and simple. Otherwise, it is difficult for users to understand the behaviors of distilled model.
    \item \textit{Compatibility:} The type of distilled model should be compatible with the original one. For instance, use of a fully-connected network to approximate a convolution neural network would lead to loss of accuracy.  
\end{itemize}
Without loss of generality, we assume the original ML model is a multi-layer convolutional neural network. Formally, given input data $\textbf{X}$, output $\textbf{Y}$, the goal of distillation is to find a matrix $\textbf{K}$ using Equation~\ref{eqn:spec}, where ``*'' denotes the matrix convolution.
\begin{equation}
\textbf{X} * \textbf{K} = \textbf{Y}
\label{eqn:spec}
\end{equation}

Since convolution is a linear-shift-invariant operation, the above regression guarantees the distilled model to be sufficiently lightweight and transparent. Under this scenario, the model computation task boils down to solving for the parameters in matrix ${\bf K}$.

\vspace{0.1in}
\textit{\textbf{Model Computation:}} To solve for ${\bf K}$, one key observation is that we can apply Fourier transformation on both sides of Equation~\ref{eqn:spec}, and by discrete convolution theorem, it gives

\begin{equation}
\begin{split}
\textbf{X} * \textbf{K} &= \textbf{Y}\\
\mathscr{F}(\textbf{X} * \textbf{K}) &= \mathscr{F}(\textbf{Y})\\
\mathscr{F}(\textbf{X}) \circ \mathscr{F}(\textbf{K}) &= \mathscr{F}(\textbf{Y})
\end{split}
\label{eqn:model}
\end{equation}
where $\circ$ is the Hadamard product. Therefore, the solution is given by this formula:
\vspace{-0.05in}
\begin{equation}
\textbf{K} = \mathscr{F}^{-1}(\mathscr{F}(\textbf{Y})/\mathscr{F}(\textbf{X}))
\label{eqn:hadamard}
\vspace{-0.05in}
\end{equation}

\textit{\textbf{Outcome Interpretation:}} The primary goal of explainable AI is to measure how each input feature contributes to the output value. Once $\textbf{K}$ is obtained, the contribution of each feature can be viewed in an indirect way -- consider a scenario where we remove this component from the original input, and let it pass through the distilled model again to produce a ``perturbed'' result. Then by calculating the difference between the original and newly generated outputs, the impact of the key feature on the output can be quantified. The intuition behind the assumption is that hiding important features are more likely to cause considerable changes to the model output. 
Formally, assume that the input is $\textbf{X} = [\textbf{x}_1, \textbf{x}_2, ...,\textbf{x}_{i-1}, \textbf{x}_{i}, \textbf{x}_{i+1}..., \textbf{x}_d]$. We define the contribution factor of $\textbf{x}_i$ as
\begin{equation}
\begin{split}
con(\textbf{x}_i) \triangleq \textbf{Y} - \textbf{X}^\prime * \textbf{K}
\end{split}
\label{eqn:contribution}
\vspace{-0.4in}
\end{equation}
where $\textbf{X}^\prime = [\textbf{x}_1, \textbf{x}_2, ...,\textbf{x}_{i-1}, \textbf{0}, \textbf{x}_{i+1}..., \textbf{x}_d]$, which is nothing but removing the target component from the original input.

As we can see, the original model distillation task has been converted into a matrix computation problem, which consists of matrix convolution, point-wise division, and Fourier transform. The first two types of operations can be inherently accelerated by GPU or TPU's built-in structure~\cite{intro}. The details for accelerating Fourier transform is described in Section~\ref{decomp}.

\subsection{Transformation of Shapley Analysis}
\label{transSP}
The transformation of Shapley value computation has been well studied in ~\cite{wang2019matrix}. This paper represents the value function as a pseudo-Boolean function, and a corresponding formula is obtained to calculate the Shapley value for graphical cooperate games. Specifically, for every pseudo logical value function $v$, we need to find its structure vector $C_v \in \mathbb{R}^{2^n} $, such that the  value equation can be  expressed into its matrix form as
\[
v(S) = C_v x^S
\]
Then according to the definition of Shapley value
\[
\phi_i = \sum\limits_{S \subseteq N/\{i\} } \frac{|S|!(M - |S| -1)!}{M!} [f_x(S \cup \{i\}) -f_x(S)]    
\]
We have 
\[
\begin{split}
    & v(S \cup \{i\}) -v(S) \\
= & C_v (x_1^S \ldots x^S_{i-1} \begin{bsmallmatrix*}1\\0 \end{bsmallmatrix*} x^S_{i+1}\ldots x^S_n - x^S_1\ldots x^S_{i-1}\begin{bsmallmatrix*}0\\1 \end{bsmallmatrix*} x^S_{i+1}\ldots x^S_n)
\end{split}
\]

Now, we can use TPU to solve these system of equations.

\subsection{Transformation of Integrated Gradients}
\label{transIG}
The computation of integrated gradient is straightforward using Equation~\ref{eq:1}. However, in actual application, the output function $F$ is usually too complicated to have an analytically solvable integral. In this paper, we address this challenge using two strategies. First, we apply numerical integration with polynomial interpolation to approximate the integral. Second, we perform interpolation using the \textit{Vandermonde matrix} to accommodate it to TPU.

The numerical integration is computed through the trapezoidal rule. Formally, the trapezoidal rule works by approximating the region under the graph of the function $F(x)$ as a trapezoid and calculating its area to approach the definite integral, which is actually the result obtained by averaging the left and right Riemann sums. The interpolation improves  approximation by partitioning the integration interval, applying the trapezoidal rule to each sub-interval, and summing the results. Let $\{x_k\}$ be the partition of $[a,b]$ such that $a < x_1 <x_2 < ...x_{N-1} <x_N < b$ and let $\Delta x_k$ be the length of the $k$-th sub-interval, then
\[
\int\limits^b_a F(x)dx \approx \sum\limits^N_{k=1} \frac{F(x_{k-1} + F(x_k))}{2} \Delta x_k
\]

Now we shift our focus on interpolation using the \textit{Vandermonde matrix} to accommodate it to TPU. The basic procedure to determine the coefficients $a_0,a_1,...,a_n$ of a polynomial 
\[P_n(x) = a_0 + a_1 x + a_2x^2 + ... a_n x^n\]
such that it interpolates the $n+1$ points
\[
(x_0,y_0), (x_1,y_1), (x_2,y_2), ..., (x_n,y_n)
\]
is to write a linear system as follows:
\[
\begin{split}
P_n(x_0)  = y_0 &\Rightarrow{}  a_0 + a_1 x_0 + a_2x_0^2 + ... a_n x_0^n  = y_0\\
P_n(x_1)  = y_1 &\Rightarrow{}  a_0 + a_1 x_1 + a_2x_1^2 + ... a_n x_1^n = y_1\\
P_n(x_2)  = y_2 &\Rightarrow{}  a_0 + a_1 x_2 + a_2x_2^2 + ... a_n x_2^n = y_2\\
 \vdots\\
P_n(x_n)  = y_n &\Rightarrow{}  a_0 + a_1 x_n + a_2x_n^2 + ... a_n x_n^n = y_n
\end{split}
\]
or, in matrix form:
\[
\begin{bmatrix}
1 & x_0 & x_0^2 & \ldots & x_0^{n-1} & x_0^n \\
1 & x_1 & x_1^2 & \ldots & x_1^{n-1} & x_1^n \\
1 & x_{n-1} & x_{n-1}^2 & \ldots & x_{n-1}^{n-1} & x_{n-1}^n \\
\vdots & \vdots & \vdots & \vdots & \vdots & \vdots\\
1 & x_n & x_n^2 & \ldots & x_n^{n-1} & x_n^n
\end{bmatrix}
\begin{bmatrix}
a_0\\
a_1\\
\vdots\\
a_{n-1}\\
a_n
\end{bmatrix}
=
\begin{bmatrix}
y_0\\
y_1\\
\vdots\\
y_{n-1}\\
y_n
\end{bmatrix}
\]
The left matrix $V$ is called as a Vandermonde matrix. We can notice that $V$ is non-singular, thus we can solve the system using TPU to obtain the coefficients.

\subsection{Data Decomposition in Fourier Transform}
\label{decomp}
In this section, we discuss how to apply data decomposition to disentangle Fourier transform computation, and further utilize computation resources to significantly accelerate the computing process. The general form of a 2-D Discrete Fourier Transform (DFT) applied on an $M\times N$ signal is defined as:
\vspace{-0.05in}
\begin{equation} \label{eq:1}
X[k,l] = \frac{1}{\sqrt{MN}}\sum_{n=0}^{N-1} \left[ \sum_{m=0}^{M-1} x[m,n] e^{-j2\pi\frac{mk}{M}} \right] e^{-j2\pi\frac{nl}{N}}
\vspace{-0.05in}
\end{equation}
where $k = 0, ..., M-1$ , $l = 0, ..., N-1$. 

\noindent If we define intermediate signal $X'$ such that
\vspace{-0.05in}
\begin{equation} \label{eq:2}
X'[k,n] \triangleq \frac{1}{\sqrt{M}} \sum_{m=0}^{M-1} x[m,n] e^{-j2\pi\frac{mk}{M}} 
\vspace{-0.05in}
\end{equation}
and plug into Equation~\ref{eq:1}, we have
\vspace{-0.05in}
\begin{equation} \label{eq:3}
X[k,l] = \frac{1}{\sqrt{N}}\sum_{n=0}^{N-1}X'[k,n] e^{-j2\pi\frac{nl}{N}}
\vspace{-0.05in}
\end{equation}

Notice the similarity between Equation~\ref{eq:2} and the definition of 1-D Fourier transform applied on a $M$-length vector:
\vspace{-0.05in}
\begin{equation}
X[k] = \frac{1}{\sqrt{M}} \sum_{m=0}^{M-1} x[m] e^{-j2\pi\frac{mk}{M}} 
\vspace{-0.05in}
\end{equation}

If we treat $n$ as a fixed parameter, then application of Equation~\ref{eq:2} is equivalent to performing a 1-D Fourier transform on the $n$-th column of the original input $M \times N$ matrix. Note that for 1-D Fourier transform, it can always be written as a product of input vector and  Fourier transform matrix. Therefore, 
we can rewrite Equation~\ref{eq:2} as: 
\vspace{-0.05in}
\begin{equation}
X'[k,n]={\bf W}_M \cdot x[m,n]
\vspace{-0.05in}
\end{equation}
where ${\bf W}_M$ is the $M \times M$ Fourier transform matrix. By varying $n$ from 1 to $N-1$ and showing results side by side, we get:
\vspace{-0.05in}
\begin{equation}    
\label{eq:6}
X'=\left[X'[k,0],\cdots,X'[k,N-1]\right] = {\bf W}_M \cdot x
\vspace{-0.05in}
\end{equation}

If we treat $k$ as a parameter and view the definition of $X'[k,n]$ as the 1-D Fourier transform with respect to the $k$-th row of input $x$, a similar expression can be obtained using the above derivation steps as:
\vspace{-0.1in}
\begin{equation}
X = X' \cdot {\bf W}_N 
\vspace{-0.05in}
\end{equation} 
where ${\bf W}_N$ is the $N \times N$ Fourier transform matrix. Using Equation~\ref{eq:6}, the final expression of X can be written as:
\vspace{-0.05in}
\begin{equation}
X = ({\bf W}_M\cdot x) \cdot {\bf W}_N 
\vspace{-0.05in}
\end{equation}

This transformed expression indicates that a 2-D Fourier transform can be achieved in a two-stage manner. First, transform all the rows of $x$ to obtain intermediate result $X'$. Second, transform all the columns of the resulting matrix $X'$. An important observation is that the required computation for each row (column) is completely independent. This implies that we can always split the computation process into sub-threads. Given $p$ individual cores involved and a $M \times N$ matrix as input, every core is assigned at most $\frac{max\{M,N\}}{p}$ 1-D Fourier transforms workload and can execute in parallel.
Our analysis reveals that merging the results afterwards exactly matches the desired 2-D Fourier transform result. Algorithm~\ref{alg:alg1} outlines the major steps in data decomposition.

\begin{algorithm}[h]
\SetKwInOut{Input}{Input}
\SetKwInOut{Output}{Output}
\Input{$M \times N$ matrix $x$, number of cores $p$}
\Output{2D Fourier transform result $X$} 
 Initialize each core $c_1, c_2,...c_p$\\
 $X = {\bf 0}$\\
 \For{each $i \in [0, ..., p-1]$}{
 Split $M/p$ rows $x_i$ from $x$\\
 $X'_i$= $Execute(c_i, x_i)$
 }
 Merge Results: $X' = [X'_1, X'_2, ..., X'_{p}]^T$\\
 \For{each $j \in [0, ..., p-1]$}{
 Split $N/p$ columns $x'_j$ from $X'$\\
 $X_j$= $Execute(c_j, x'_j)$
 }
 Merge Results: $X = [X_1, X_2, ..., X_{p}]$\\
 \textbf{return} $X$\\
 $   $\\
 \textbf{procedure} $Execute(c_i, x_i)$\\
   $res = {\bf 0}$ \\
   \For{each $r \in x_i$}{
   $r'= \mathscr{F}(r)$\\
   }
   $res = merge(res, r)$\\
   \textbf{return} $res$\\
 \textbf{end procedure}\\

\caption{Acceleration of Fourier Transform}
\label{alg:alg1}
\end{algorithm}

\subsection{Parallel Computation of Multiple Inputs}
\label{paral}
In addition to exploiting hardware to accelerate Fourier transform, we make use of parallel computing to further improve the time efficiency. Notice in the training phase, multiple inputs will be fed into the model to generate corresponding outputs. The above data decomposition technique is applied on each individual input such that the computation cost is distributed among several cores. Extending from single to multiple input is simple and only requires one-step further utilization of parallel computation.

\begin{figure}[htbp]
\centering
\vspace{-0.1in}
\includegraphics[scale=0.3]{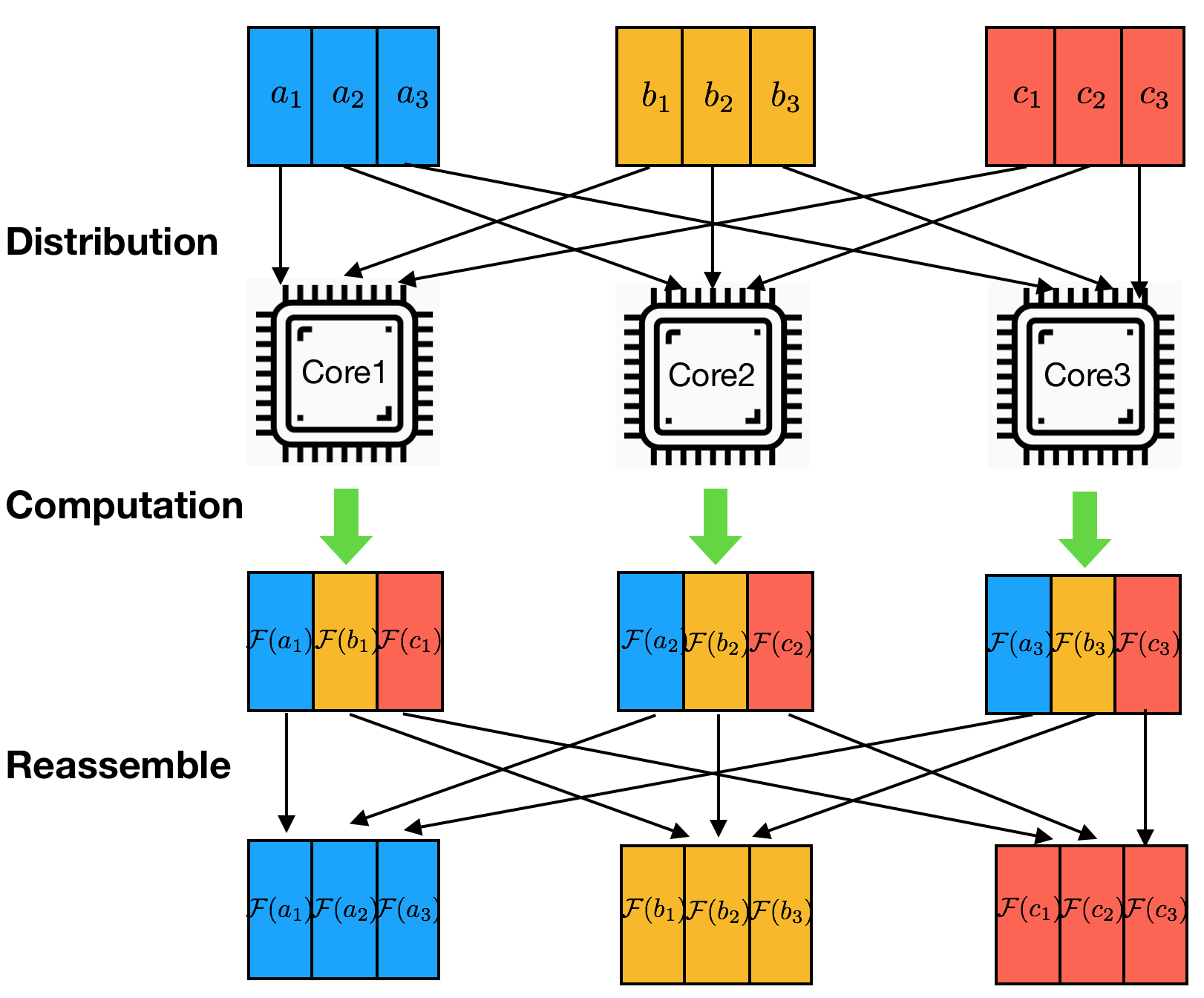}
\vspace{-0.05in}
\caption{An example of parallel computation in our framework. Each input is separated into pieces for  multiple cores to run in parallel. The outputs are reassembled to obtain the results.}
\label{fig:parallel}
\vspace{-0.1in}
\end{figure}

 An illustrative example is shown in Figure~\ref{fig:parallel} where the goal is to perform 1-D Fourier transform on each column of three input matrices. First, each input matrix is segmented into pieces and each core obtains a slice of them. Next, each piece is assigned to an individual core to perform the Fourier transform. During computation, an internal  table is utilized to keep track of the distribution to guide the process of reassembling. 
In terms of matrix multiplication, the framework is exactly the same except the fact that \textit{block matrix multiplication} is applied. Original matrices are partitioned into small blocks. By performing multiplication between blocks and merging afterwards, we achieve same-level of parallel computing efficiency. Due to the data decomposition step  applied in Section~\ref{decomp}, the whole computing procedure contains Fourier transform, matrix multiplication and point-wise division only, which indicates parallelism is maintained across the entire computation process.
 
 In this work, the communication among separate cores is implemented with $tf.cross\_replica\_sum$ and is required at every iteration of reassembly process to compute the summation of the partial matrices across the cores. The proposed data decomposition and the parallel implementation not only efficiently utilizes GPU/TPU’s strength in matrix multiplication but also requires minimal communication time, which leads to drastic improvement in acceleration performance.
\section{Experimental Evaluation}\label{exp}
This section evaluates the effectiveness of our proposed framework in accelerating explainable AI. We first describe the experimental setup. Next, we present acceleration results.

\subsection{Experimental Setup}

\begin{table*}[h]
    \centering
    \caption{Comparison of accuracy and classification time for various benchmarks.}
    \vspace{-0.05in}
    \label{tb_1}
    \begin{tabular}{|c|c|c|c|c|c|c|c|c|c|c|c|}
        \hline
        & \multicolumn{3}{c|}{CPU-based Acceleration} 
        & \multicolumn{3}{c|}{GPU-based Acceleration} 
        & \multicolumn{5}{c|}{TPU-based Acceleration}\\
        \hline
        Benchmark  & Accuracy& Training- & Testing-  & Accuracy & Training- & Testing- & Accuracy & Training- & Testing- & Speedup. & Speedup.\\
        &  (\%) &  time(s) & time(s) &  (\%) & time(s) & time(s) & (\%) & time(s) & time(s) & /CPU & /GPU\\
        \hline
        VGG19  & 94.06 & 24.2 & 10.9  & 92.08 & 0.25 & 0.08  & 96.37 & 0.4 & 0.14 & 65x & 0.61x\\
        \hline
        ResNet50  & 78.99 & 176.2 & 129.8 &  86.87 & 19.1 & 9.4  & 87.47 & 4.3 & 2.6 & 44.5x & 4.13x\\
        \hline
        {\bf Average} &  {\bf 86.52} & {\bf 100.2} & {\bf 70.35} & {\bf 89.47} & {\bf 9.67} & {\bf 4.84} &  {\bf 91.92} & {\bf 2.35} & {\bf 1.37} & {\bf 54.7x} & {\bf 3.9x}\\
        \hline
    \end{tabular}
    \vspace{-0.15in}
\end{table*}

The experiments were conducted on a host machine with Intel i7 3.70GHz CPU, equipped with an external NVIDIA GeForce RTX 2080 Ti GPU, which is considered as the state-of-the-art GPU accelerator for  ML algorithms. We also utilize Google's Colab platform to access Google Cloud TPU service. In our evaluation, we used TPUv2 with 64 GB High Bandwidth Memory (HBM), and 128 TPU cores. We developed code using Python~\cite{Python} for model training and PyTorch 1.6.0~\cite{Pytorch} as the machine learning library. We have used the following two populor benchmarks in our study.

\begin{enumerate}
    \item The VGG19~\cite{VGG19} classifier for \textit{CIFAR-100}~\cite{CIFAR} image classification.
    \item The ResNet50~\cite{ResNet} network for \textit{MIRAI}~\cite{MalwareDataset} malware detection.
\end{enumerate}

We have used the following three hardware configurations to highlight the importance of our proposed hardware acceleration approach. To address the compatibility of proposed optimization approach (Algorithm~\ref{alg:alg1}), all the proposed optimization methods (task transformation, data decomposition, parallel computation) are deployed on all 3 accelerators:
\begin{enumerate}
\item \textbf{CPU}: Traditional execution in software, which is considered as baseline method.
\item \textbf{GPU}: NVIDIA GeForce 2080 Ti GPU, which is considered as state-of-the-art ML acceleration component.
\item \textbf{TPU}: Google's cloud TPU, a specific ASIC designed to accelerate machine learning procedure.
\end{enumerate}

We first evaluated classification performance by reporting ML models' classification accuracy and execution time. Next, we evaluate our proposed framework's energy performance by measuring its performance-per-watt on each hardware, and further record its power consumption under different workload. Then we apply all three XAI methods to explain the model's output, and report the average time for completing outcome interpretation step for each configuration. Finally, we discuss the effectiveness of proposed method in interpreting classification results.

\subsection{Comparison of Accuracy and Classification Time}
Table~\ref{tb_1} compares the classification time and accuracy. Each row represents a specific model structure trained with corresponding hardware configuration. For both training time and testing time, each entry represents time cost of 10 epochs on average. As we can see, with sufficient number of training epochs, all methods obtain reasonable classification accuracy. However, when it comes to time-efficiency, the CPU-based baseline implementation lags far behind the other two, which achieved the slowest speed. On VGG19, GPU provides the best acceleration performance, which provides 65x speedup compared to the baseline implementation. This clearly indicates the great compatability between hardware accelerator and our proposed framework. In case of ResNet50, an even higher speedup was obtained by TPU, showing its acceleration potential in large scale neural networks by providing around four times speedup than GPU. The drastic improvement (44.5x) compared to the baseline method also leads to significant energy savings, as described in the next section. Note that  Table~\ref{tb_1} provided results in terms of model distillation. We have observed the similar trend for the other explainable AI algorithms (Shapley analysis and integrated gradients).



\subsection{Power Consumption of Different Hardware Accelerators}
\label{sec:power}

\begin{figure*}[h]
\centering
\includegraphics[scale =0.76]{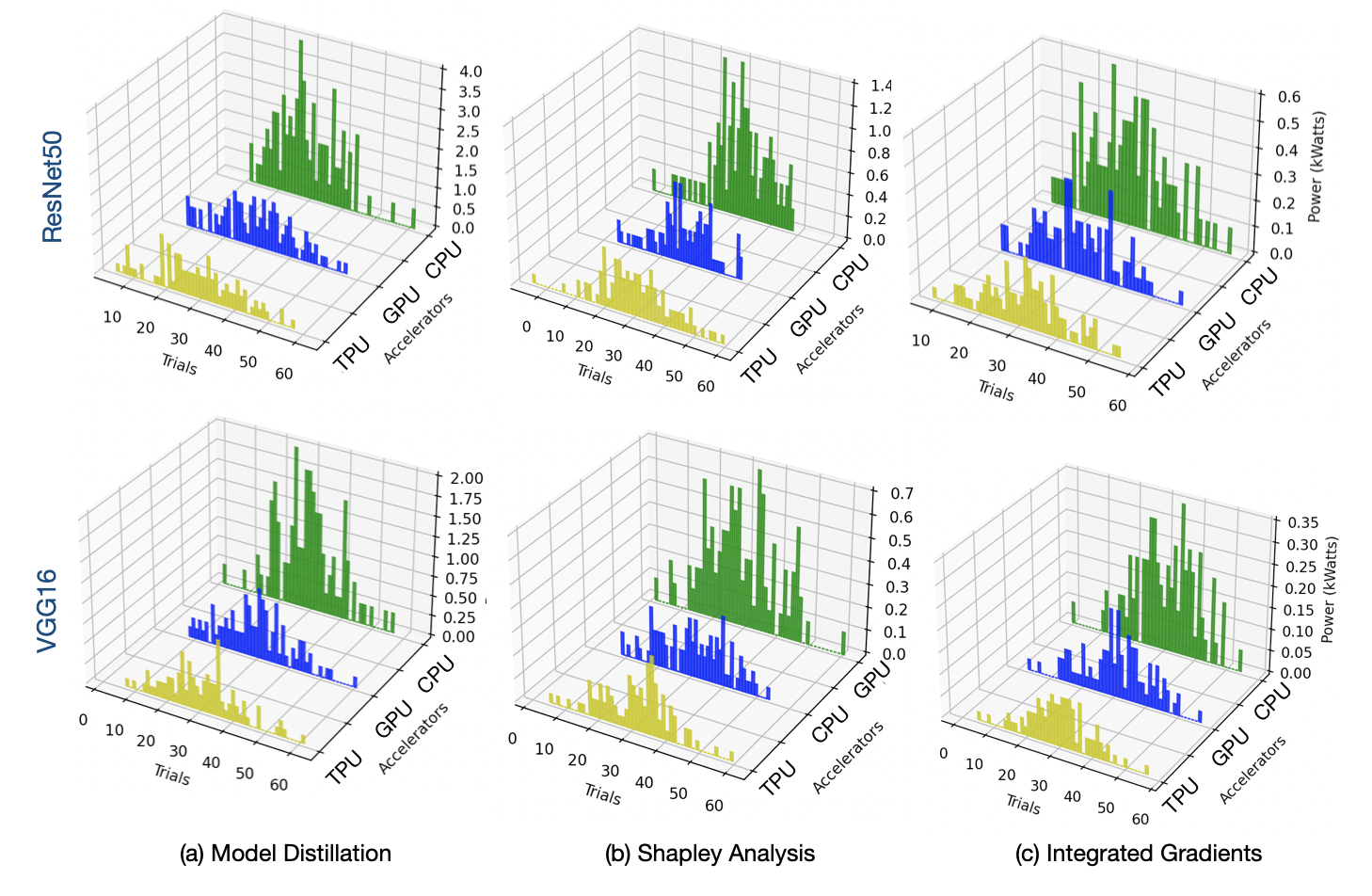}
\caption{The power consumption of different hardware accelerators across 100 tasks on two different machine learning models (ResNet50 and VGG16). The ResNet 50 model consists of 50 layers consisting of $>$1000 nodes, while the VGG16 model is composed with layers in a depth of 16 and 138 trainable parameters. We plot the detailed power performance across 60 trials for three XAI algorithms (a) model distillation, (b) Shapley analysis, and (c) integrated gradients.}
\label{fig:hists} 
\vspace{-0.1in}
\end{figure*}

We have evaluated the power consumption of three configurations (CPU, GPU and TPU) under different workloads, as shown in Figure~\ref{fig:hists}. We measure the power consumption in kWatts for the three hardware accelerators on two different types of ML models (ResNet50 and VGG16). We repeat the experiment across 100 trials with various problem size, within each of them we perform all three kinds of XAI techniques (model distillation, Shapley analysis, and integrated gradients). In terms of energy efficiency, TPUs outperform the other two, which demonstrates the compatibility between TPU and our proposed method, which maximizes data decomposition to create a high-level parallel computing environment, establishing the balanced workloads across cores. Interestingly, for some special tasks, GPU can even cause more energy consumption than CPU. The bottleneck is typically caused not by computation, but by thread divergence and memory access. For tiny-scale problems, the advantage of efficient computation cannot compensate the extra cost caused by memory allocation. TPU avoid such problem since it employs integer arithmetic instead of floating-point calculations, which significantly reducing the required  hardware size and power consumption. 

\begin{figure}[h]
\vspace{-0.15in}
\centering
\includegraphics[scale =0.35]{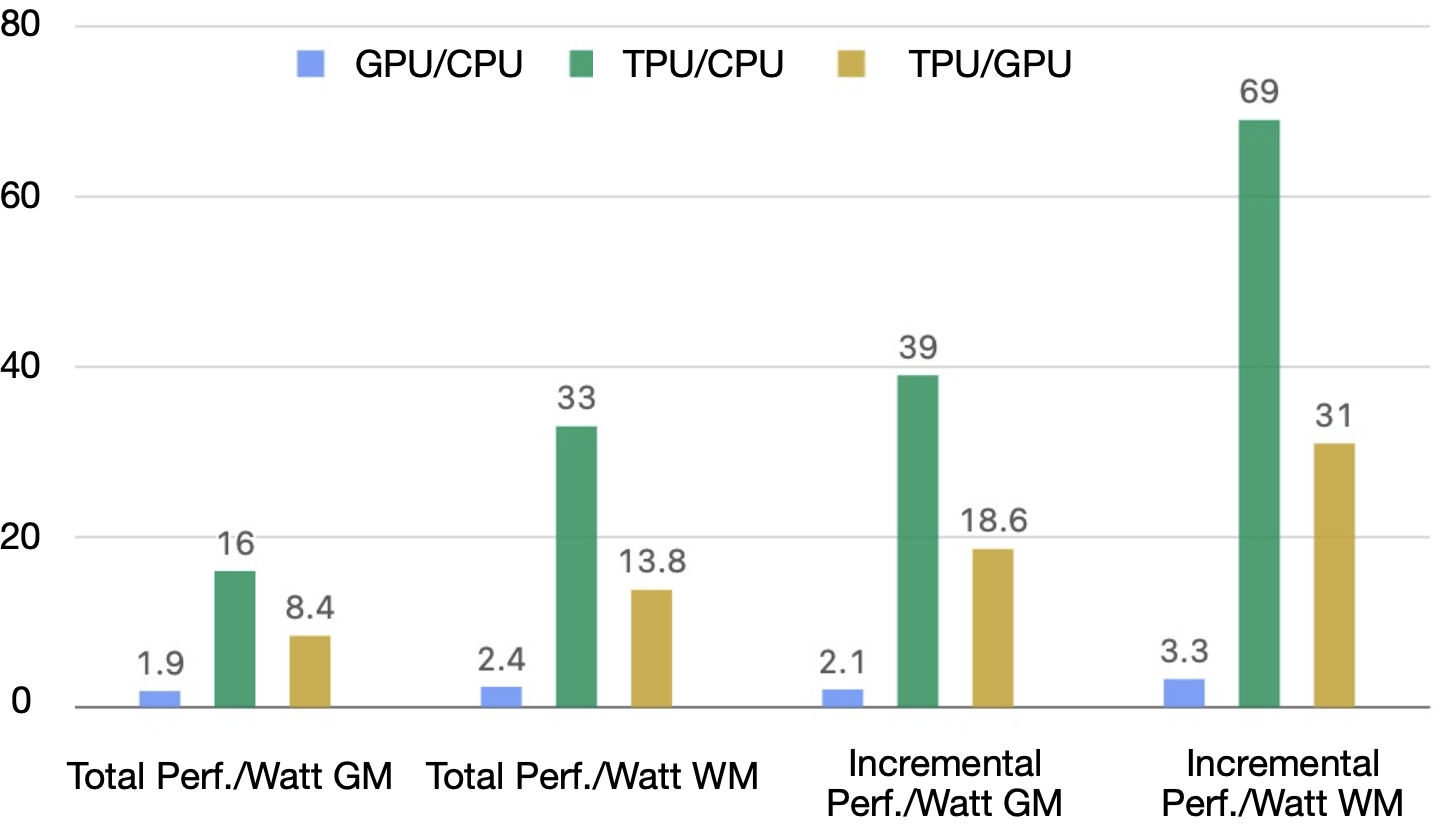}
\vspace{-0.2in}
\caption{Relative performance/Watt of GPU (blue bar) and TPU (green bar) over CPU, and TPU versus GPU (yellow bar) using model distillation. The total perf./watt includes host CPU power, while incremental ignores it. {\bf GM} and {\bf WM} are the geometric mean and weighted arithmetic mean, respectively.}
\label{fig:ppw} 
\vspace{-0.1in}
\end{figure}

We further evaluate the energy efficiency of different hardware accelerators. 
Figure~\ref{fig:ppw} shows the geometric and weighted mean performance/Watt for the RTX 2080 Ti GPU and Google's TPU relative to the CPU. Similar to~\cite{jouppi2017datacenter}, we calculate performance/Watt in two different ways. The first one (referred as `total') computes the total power consumption which consists of the power consumed by the host CPU as well as the actual execution performance/Watt for the GPU or TPU. The second one (referred as `incremental') does not consider the host CPU power, and therefore, it reflects the actual power consumption of the GPU or TPU during acceleration. As we can see from Figure~\ref{fig:ppw}, for total-performance/watt, the GPU implementation is 1.9X and 2.4X better than baseline CPU for geometric mean (GM) and weighted arithmetic mean (WM), respectively. TPU outperforms both CPU (16x on GM and 33X on WM) and GPU (8.4X on GM and 13.8x on WM) in terms of total performance/watt. For incremental-performance/watt, when host CPU's power is omitted, the TPU shows its dominance in energy efficiency over both CPU (39x on GM and 69X on WM) and GPU (18.6x on GM and 31X on WM). 

In terms of energy efficiency, GPU-based acceleration is better than software execution (CPU) while TPU outperforms both GPU and CPU. Our proposed method fully utilizes data decomposition to create a high-level parallel computing environment where both GPU an TPU benefit from it to balance the workloads on every single core. Although both GPU and TPU have the advantage of utilizing parallel computing to fulfill the proposed framework, TPU provides better performance/watt primarily due to the `quantification' property of TPU. Quantification is a powerful mechanism to reduce the cost of neural network prediction as well as the reduction in memory. As mentioned in Section~\ref{sec:gputpu}, the use of integers instead of floating point calculations greatly reduces the hardware size and power consumption of the TPU. Specifically,  TPU can perform 65,536 8-bit integer multiplications in a cycle, while mainstream GPUs can perform few thousands of 32-bit floating-point multiplications. As long as 8 bits can meet the accuracy requirements, it can bring significant performance improvement. While both TPU and GPU based acceleration can achieve fast explainable machine learning, the TPU-based implementation is the best in terms of energy efficiency.


\subsection{Time Efficiency of Outcome Interpretation}
In this section, we demonstrate the time efficiency of proposed method on explaining ML models.
The average time for performing outcome interpretation of three considered XAI algorithms for every 10 input-output pairs is presented in Table~\ref{tb_2}, Table~\ref{tb_3}, and Table~\ref{tb_4}, respectively. The VGG19 result demonstrates that the TPU method obtains the best result as it is 36.2x and 1.9x faster than CPU and GPU based implementations, respectively. As expected, the improvements are even higher in case of ResNet50, where 39.5x and 4.78x speedup obtained over CPU and GPU based approaches, respectively. Since the outcome interpretation procedure can be completed in few seconds, our proposed acceleration makes it possible to embed explainable ML during model training in diverse applications.
Note that model distillation takes longer time since it maintains a separate model, and needs to train two models. Irrespective of the XAI model, TPU provides the best performance in terms of time.

\begin{table}[ht]
    \centering
    \caption{Average time (seconds) for outcome interpretation using Model Distillation} 
        \vspace{-0.05in}
    \begin{tabular}{|c|c|c|c|c|c|}
        \hline
        Model & CPU & GPU  & TPU & Impro./CPU & Impro./GPU \\
        \hline
        VGG19 & 550.7 & 29  & 15.2s & 36.2x & 1.9x\\
        \hline
        ResNet50 & 1456.1 & 176  & 36.8s & 39.5x & 4.78x\\
        \hline
        {\bf Average} & {\bf 1003.4} &{\bf 102.5}  & {\bf 26.0} & {\bf 38.6x} & {\bf 3.94x} \\
        \hline
    \end{tabular}
    \label{tb_2}
\end{table}

\begin{table}[ht]
    \centering
    \caption{Average time (seconds) for outcome interpretation using Shapley Values} 
        \vspace{-0.05in}
    \begin{tabular}{|c|c|c|c|c|c|}
        \hline
        Model & CPU & GPU  & TPU & Impro./CPU & Impro./GPU \\
        \hline
        VGG19 & 580.2 & 77.1  & 18.3 & 16x & 3x\\
        \hline
        ResNet50 & 50.1 & 11.6  & 3.8s & 4.5x & 3.2x\\
        \hline
        {\bf Average} & {\bf 365.4} &{\bf 44.5}  & {\bf 11.6} & {\bf 5.8x} & {\bf 2.4x} \\
        \hline
    \end{tabular}
    \label{tb_3}
\end{table}

\begin{table}[ht]
    \centering
    \caption{Average time (seconds) for outcome interpretation using Integrated Gradients} 
        \vspace{-0.05in}
    \begin{tabular}{|c|c|c|c|c|c|}
        \hline
        Model & CPU & GPU  & TPU & Impro./CPU & Impro./GPU \\
        \hline
        VGG19 & 443.1 & 17.2  & 4.5 & 25.7x & 3.8x\\
        \hline
        ResNet50 & 17.3 & 1.6  & 0.8 & 10.8x & 2x\\
        \hline
        {\bf Average} & {\bf 230.2} &{\bf 9.4}  & {\bf 2.65} & {\bf 18.2x} & {\bf 2.9x} \\
        \hline
    \end{tabular}
    \label{tb_4}
\end{table}

\begin{figure}[htbp]
\vspace{-0.1in}
\centering
\includegraphics[scale=0.29]{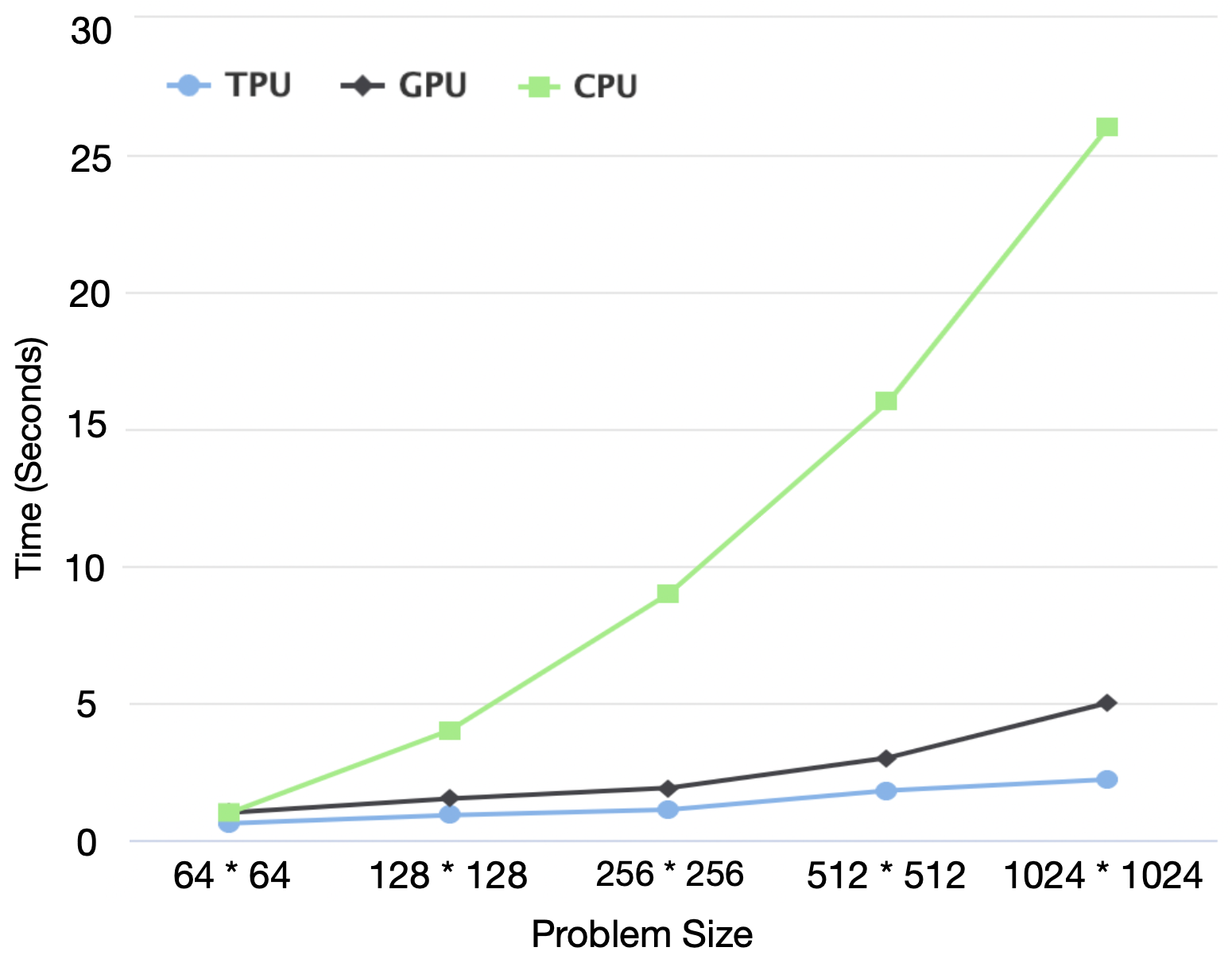}
\vspace{-0.1in}
\caption{Scalability of three acceleration methods for model distillation. In this study, we consider  ML problems with approximately one hundred tasks at a time.}
\label{fig:sz}
\vspace{-0.2in}
\end{figure}

To demonstrate the scalability of our proposed method, we have randomly selected several matrices with varying sizes and compared the time efficiency as shown in Figure~\ref{fig:sz}. It is expected that the time will increase with the increase in the size of the matrices. Figure~\ref{fig:sz} shows that our proposed implementations on GPU and TPU based architectures are  scalable across various problem sizes. There are two reasons for the scalability of our approach. (1) Our proposed approach utilizes data decomposition technique to break larger matrices into smaller sub-matrices. (2) Another dimension of improvement comes from the fact that these smaller sub-matrices are distributed across multiple MXU inside each individual core. This drastically reduces the bandwidth requirement during the computation and leads to a significant improvement in computation speed. For matrices in the size of 1024 × 1024, TPU implementation is more than 30x faster than the baseline one. This indicates that for training and outcome interpretation on large-scale neural networks (with tens of thousands of matrix operations), our proposed TPU-based acceleration can save hours of computation time, which also leads to significant energy savings.
\color{black}

\subsection{Outcome Interpretation Examples}
Since we are using hardware components for accelerating explainable ML, our method not only achieved faster classification, but also provides effective explanation of classification results. We have evaluated outcome interpretation in a wide variety of scenarios. We first provide a simple example of outcome interpretation to highlight the similarity of the three explainable AI models (model distillation, Shapley analysis, and integrated gradients) that consider different features in final classification. Next, we provide examples to illustrate their fundamental differences in computing the importance of different features.

\begin{figure}[h]
\vspace{-0.15in}
\centering
\includegraphics[scale =0.35]{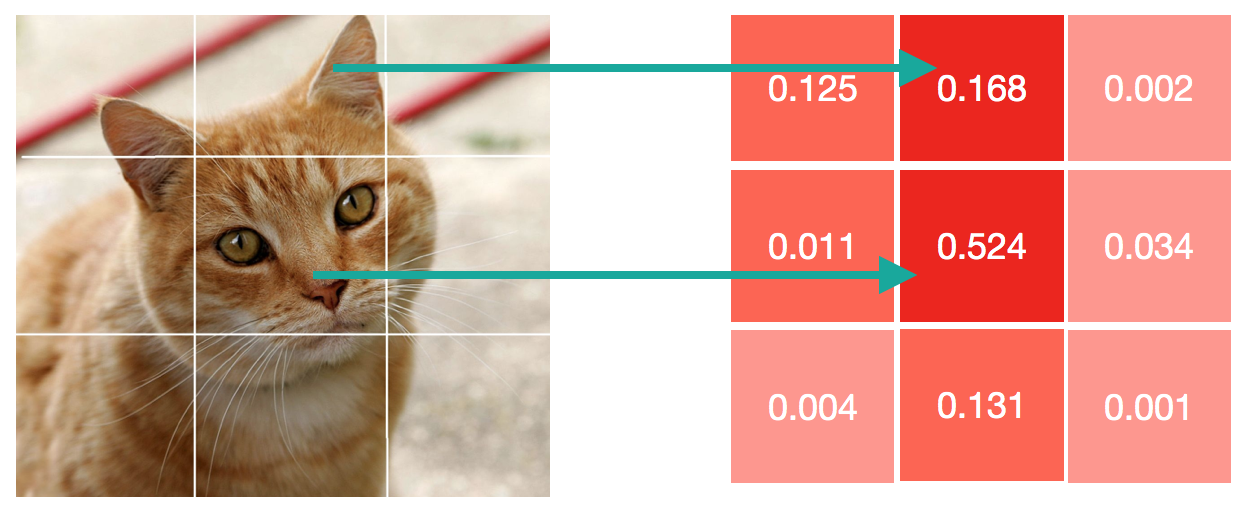}
\vspace{-0.1in}
\caption{Interpretation of CIFAR image's classification}
\vspace{-0.1in}
\label{fig:interp2} 
\end{figure}

Figure~\ref{fig:interp2} shows an example of interpreting the classification results for a picture from CIFAR-100 dataset. We segmented the given image into square sub-blocks, and the explainable ML framework is applied to compute contribution factor of each individual block towards the classifier's output, so that it can illustrate what part is crucial for the classifier to distinguish it from the other categories. In the given picture, the cat's face (central block) and ear (mid-up block) are the keys to be recognized as `cat'.

\begin{figure}[h]
\vspace{-0.15in}
\centering
\includegraphics[scale =0.38]{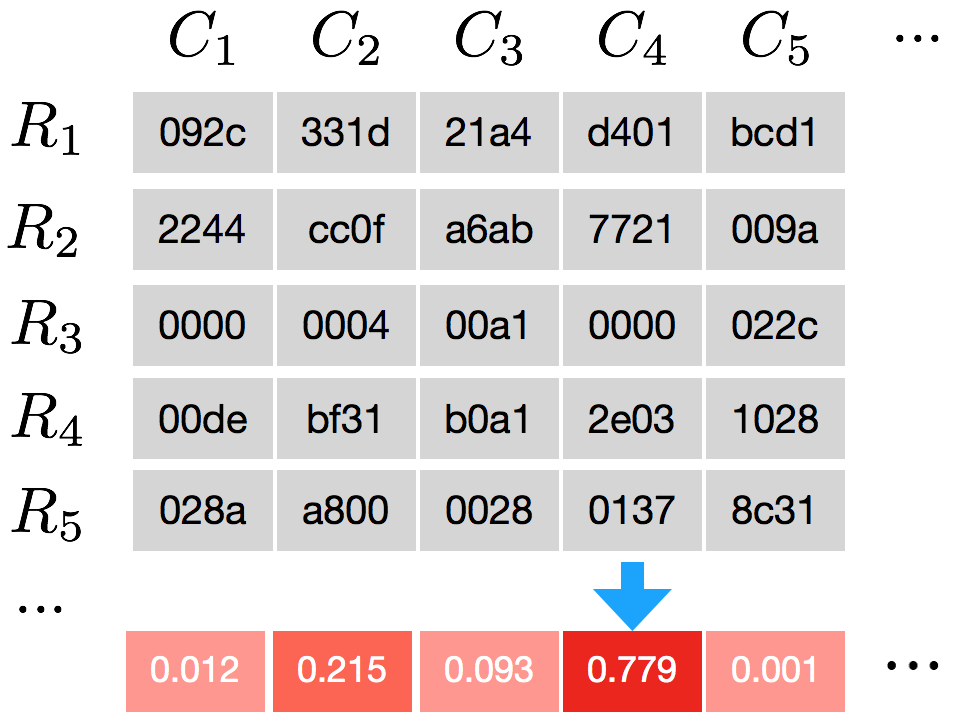}
\vspace{-0.1in}
\caption{Interpretation of MIRAI malware traced signals}
\label{fig:interp1} 
\vspace{-0.1in}
\end{figure}

\vspace{0.1in}
\noindent \underline{Outcome Interpretation using Model Distillation}
To understand how model distillation gathers insights, let us consider an example on malware detection from ResNet50. The ML-based detector receives running data of MIRAI malware \cite{MalwareDataset} as input in the format of a trace table, where each row represents the hex values in a register in specific clock cycles (each column represents a specific clock cycle). Figure~\ref{fig:interp1} shows a snapshot of the trace table.

Our proposed method computed the corresponding contribution factor of each clock cycle towards the output using model distillation. Contribution factors are shown as weights in the last (colored) row. Clearly we can see that the weight of $C_2$ is significantly larger than the others. By tracing the execution, it has been shown that  $C_2$ corresponds to the timestamp of assigning value to the variable ``\textit{ATTACK\_VECTOR}" in Mirai. This variable records the identity of attack modes, based on which the bot takes relative actions to perform either a UDP attack or DNS attack. This attack-mode flag is the most important feature of a majority of malware bot programs, and our proposed method successfully extracted it from the traces to illustrate the reason for classifying it as a malware. This interpretation not only provides confidence in malware detection but also helps in malware localization.

\begin{figure}[htbp]
\centering
\vspace{-0.1 in}
\includegraphics[scale=0.52]{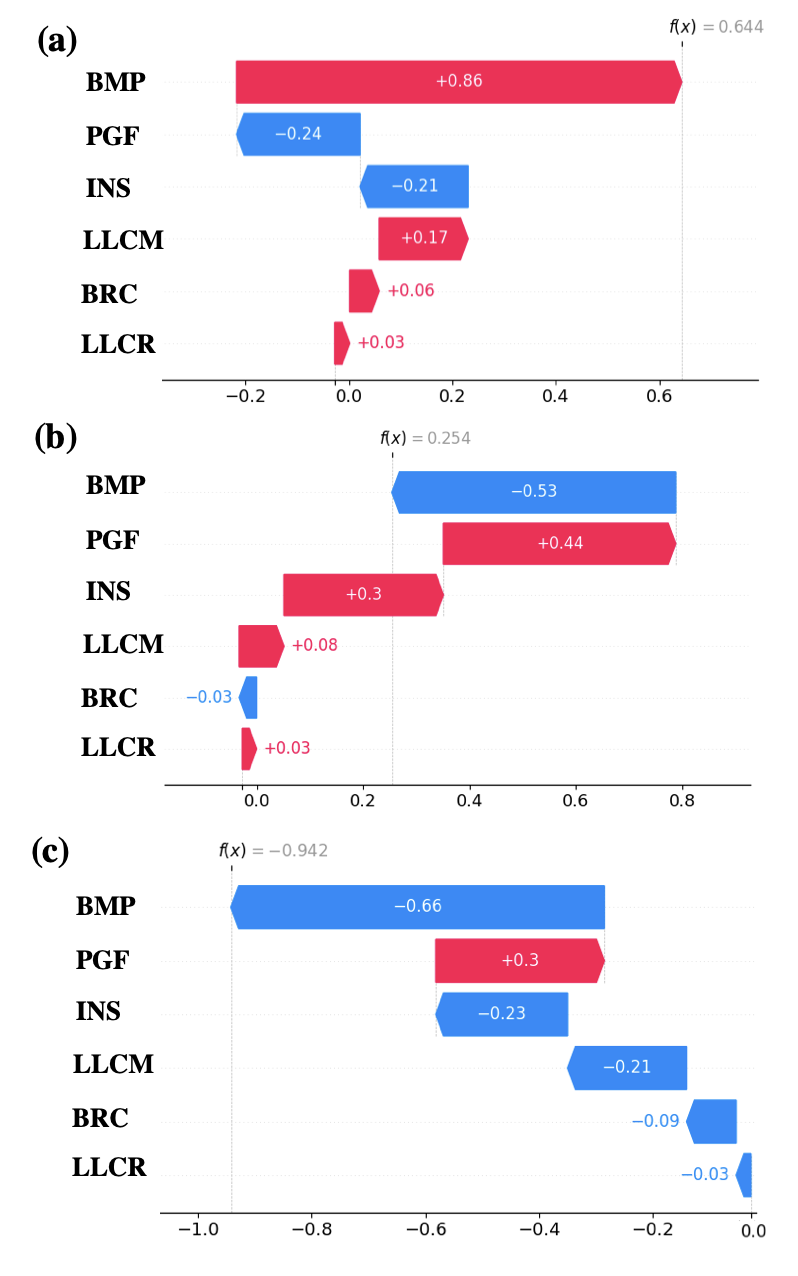}
\vspace{-0.2in}
\caption{The SHAP values for 3 example cases. (a) A Spectre attack program example, (b) A Meltdown attack example and (c) A benign program. The SHAP values clearly illustrate the major features that lead to the model output.}
\label{fig:shaps}
\end{figure}

\vspace{0.1in}
\noindent \underline{Outcome Interpretation using Shapley Analysis}:
Next, we consider the outcome interpretation through Shapley value analysis for classification task. Figure~\ref{fig:shaps} shows the waterfall plot of SHAP values from 3 samples:  (a) and (b) are true positive samples, and (c) is a true negative one. In order to detect Spectre and Meltdown attacks, it considers beneficial hardware performance counters: BMP (branch mispredictions), PGF (page faults), INS (instructions), LLCM (low-level cache misses), BRC (branches), and LLCR (low-level cache references). The waterfall plot clearly demonstrates the contribution of each feature and how they affect the decision. The plus or minus sign illustrate whether the specific feature is supporting the sample to be positive (red bars), or voting for the negative (blue bars). The SHAP values along with each bar show their exact impact, and the summation of all SHAP values is compared with the threshold to give the final decision. As we can see from the figure, BMP and PGF are among the most important features. This is reasonable and matches with our analysis in Section~\ref{relwork}. Note that in both (a) and (b), the selected attack programs are adversarial samples. In (a), we intentionally cause more page faults to mess up the feature pattern, as the PGF feature provides negative contribution to the final decision. Nevertheless, the waterfall plot clearly illustrates that our proposed model is still able to assign larger weights to BMP feature so that it can correctly predict the attack. While in (b), we insert redundant non-profit loops to create more branch mispredictions. The redundant branch mispredictions introduces the biggest negative contribution for the BMP feature to model's prediction. Since the total number of instructions are also increased, the proposed model is able to produce correct prediction with the help of the clue from total instruction numbers, reflected by the positive contribution from INS.

\begin{figure}[htbp]
\centering
\vspace{-0.1 in}
\includegraphics[scale=0.36]{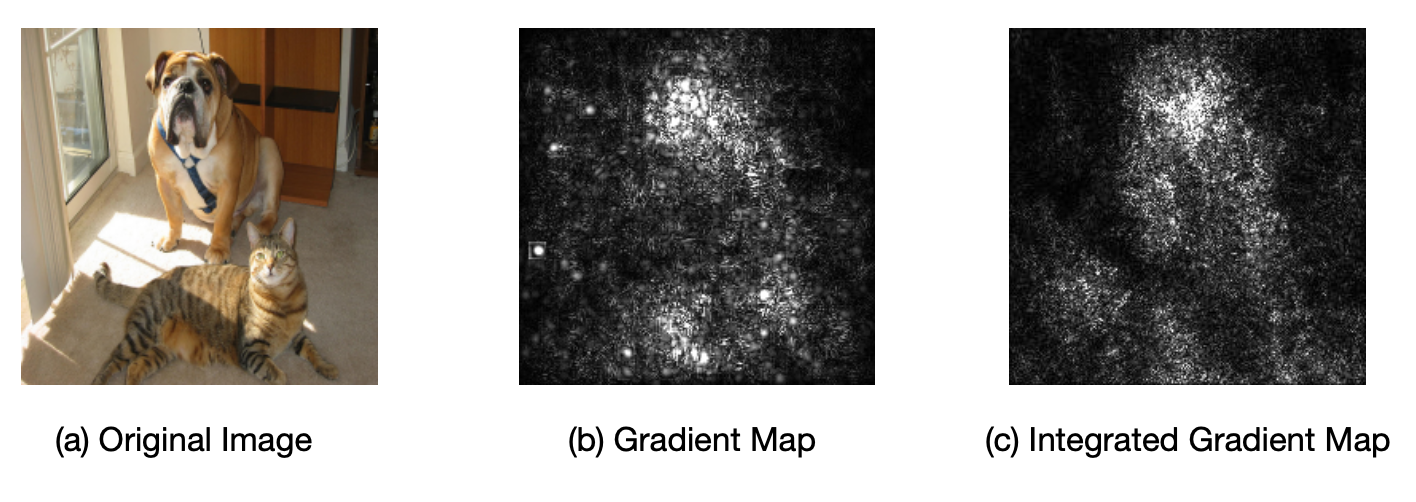}
\vspace{-0.15in}
\caption{Explainability using integrated gradients - (a) the original image, (b) the gradients map of the given image, and (c) the integrated gradients map of the given image.}
\vspace{-0.2in}
\label{fig:IGs}
\end{figure}

\vspace{0.1in}
\noindent \underline{Outcome Interpretation using Integrated Gradients}:
Finally, we consider the outcome interpretation through Integrated Gradients for an image classification example. We utilize the path-integrated gradients library provided in~\cite{goh2021understanding}, and the result is demonstrated in Figure~\ref{fig:IGs}, where we have (a) he original image (b) the gradient map of the given image, and (c) the integrated gradients map of the given image. In gradient maps ((b) \& (c)), the lightness of pixels represents their contribution towards model decision. As a result, the model's outcome can be explained by highlighting the pixels that are the most reactive and likely to quickly change the output, but only makes sense for small deviations away from the original input. As we can see from Figure~\ref{fig:IGs}(b), traditional gradient values cannot accurately reflect the distinguishing features from the input, and is easy to be disturbed by noise, inducing a random scattering of attribution values. Instead, the completeness axiom of IGs (mentioned in Section~\ref{sec:relIG}) gives a stronger (and more complete) representation of what went into the output. This is because the gradients received for saliency maps are generally in the model’s saturated region of gradients.

\section{Conclusion}\label{conclude}
While explainable AI techniques are popular, their long running time severely restrict their applicability in many domains. In this paper, we address this fundamental bottleneck using hardware-based acceleration to provide explainability (transparency) of machine learning models in a reasonable time. This paper made several major contributions. We propose an efficient mechanism to transform diverse explainable AI algorithms (model distillation, Shapley analysis, and integrated gradients) to linear algebra computations. The transformed model is able to fully exploit the inherent ability of hardware accelerators in computing ultra-fast matrix operations. Moreover, it enables parallel computing by performing data decomposition to break a large matrix into multiple small matrices. Experimental evaluation on a diverse set of benchmarks demonstrated that our approach is scalable and able to meet real-time constraints. Our studies reveal that our proposed framework can effectively utilize the inherent advantages of both TPU and GPU based architectures. Specifically, TPU-based acceleration provides drastic improvement in interpretation time (39x over CPU and 4x over GPU) as well as energy efficiency (69x over CPU and 31x over GPU) for both image classification and malware detection benchmarks.


\bibliographystyle{IEEEtran}
\bibliography{zhixin.bib}

\begin{IEEEbiography}[{\includegraphics[width=1in,clip]{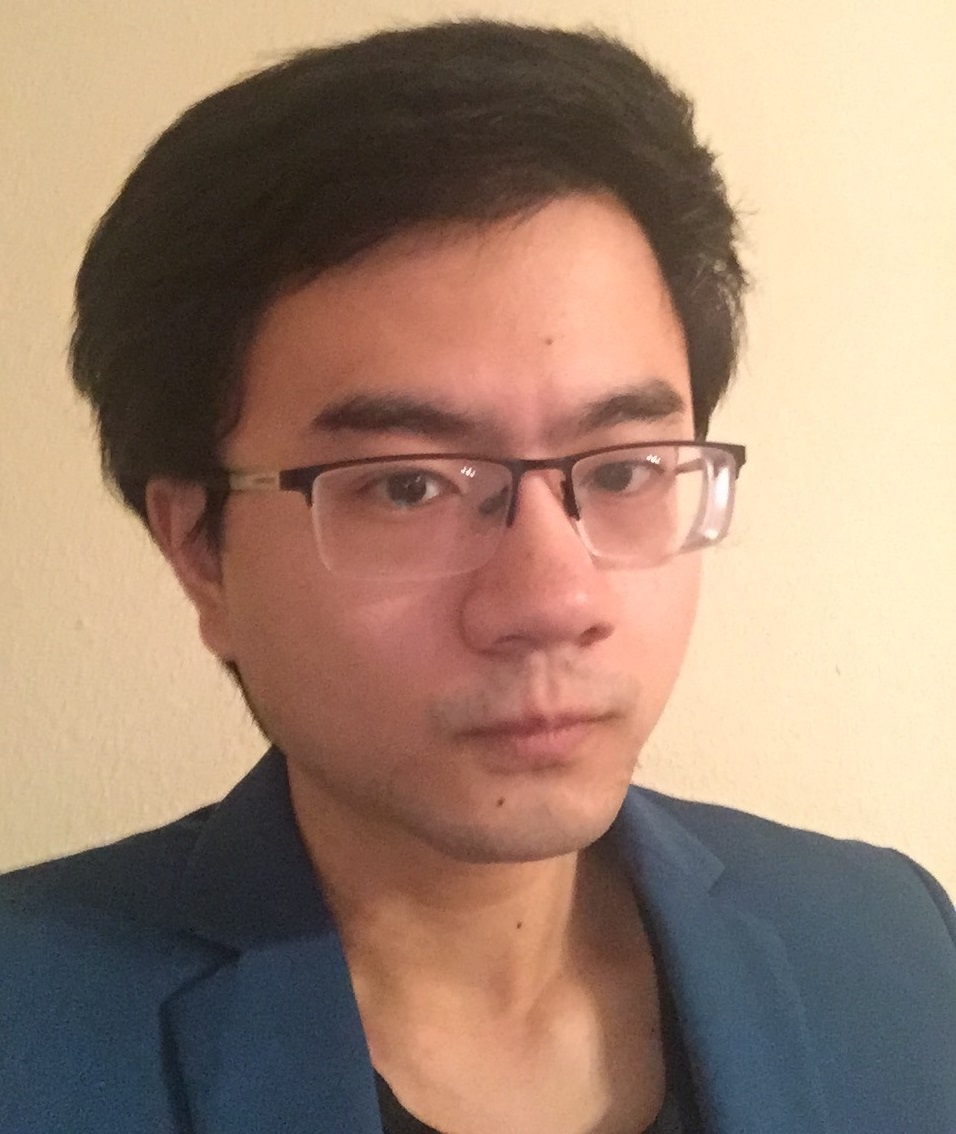}}]{Zhixin Pan} is a post-doctoral researcher in the Department of Computer \& Information Science \& Engineering at the University of Florida. He received his Ph.D. in Computer Science from the University of Florida in 2022. He received his B.E. in the Department of Software Engineering from Huazhong University of Science \& Technology, Wuhan, China in 2015. His area of research includes hardware security, post-silicon debug, data mining, quantum computing, and machine learning.
\end{IEEEbiography}

\begin{IEEEbiography}[{\includegraphics[width=1in,clip,keepaspectratio]{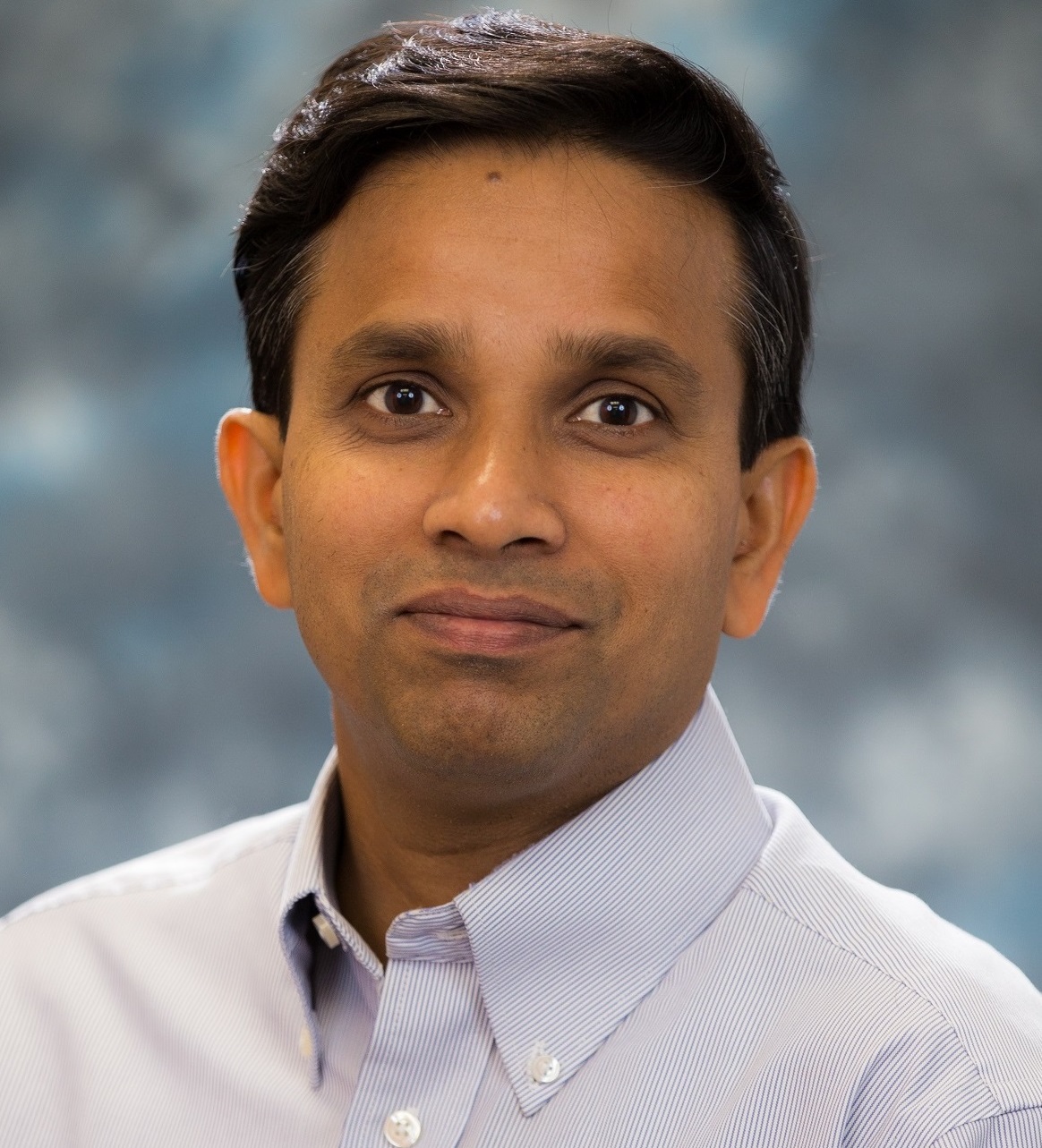}}]{Prabhat Mishra}
is a Professor in the Department of Computer and Information Science and Engineering at the University of Florida. He received his Ph.D. in Computer Science from the University of California at Irvine in 2004. His research interests include embedded systems, hardware security and trust, system-on-chip validation, machine learning, and quantum computing. He currently serves as an Associate Editor of IEEE Transactions on VLSI Systems and ACM Transactions on Embedded Computing Systems. He is an IEEE Fellow, a Fellow of the American Association for the Advancement of Science, and an ACM Distinguished Scientist.
\end{IEEEbiography}

\end{document}